%% file: root.tex
\documentclass[journal]{IEEEtran} 
\usepackage{amsmath,amsfonts}
\usepackage{array}
\usepackage{textcomp}
\usepackage{stfloats}
\usepackage{url}
\usepackage{verbatim}
\usepackage{graphicx}
\usepackage{cite}
\usepackage{svg}
\usepackage{color}
\usepackage{subfigure}
\usepackage{algorithmic}
\usepackage[ruled,vlined,linesnumbered]{algorithm2e}
\usepackage{amssymb}  
\usepackage{booktabs}
\usepackage{multirow}
\usepackage{threeparttable}
\usepackage{gensymb}
\usepackage{epstopdf}
\usepackage{hyperref}
\usepackage[normalem]{ulem}
\usepackage{CJKutf8}
\hypersetup{
    colorlinks=true,
    linkcolor=red,
    filecolor=magenta,      
    urlcolor=blue,
    citecolor=blue,
}

\begin{document}

\title{SLIM: Scalable and Lightweight LiDAR Mapping \\in Urban Environments}

\author{Zehuan Yu, Zhijian Qiao, Wenyi Liu, Huan Yin, and Shaojie Shen
\thanks{This work was supported in part by the HKUST Postgraduate Studentship, and in part by the HKUST-DJI Joint Innovation Laboratory.}
\thanks{Zehuan Yu, Zhijian Qiao, Huan Yin, and Shaojie Shen are with the Department of Electronic and Computer Engineering, The Hong Kong University of Science and Technology, Hong Kong, China. E-mail: zyuay@connect.ust.hk, zqiaoac@connect.ust.hk, eehyin@ust.hk, eeshaojie@ust.hk.}
\thanks{Wenyi Liu is with the Department of Mechanism Engineering, Hong Kong University, Hong Kong, China. Email: liuwenyi@connect.hku.hk}
\thanks{Corresponding author: Huan Yin}
}

\markboth{Accepted for Publication in IEEE Transactions on Robotics}%
{}



\maketitle

\begin{abstract}

LiDAR point cloud maps are extensively utilized on roads for robot navigation due to their high consistency. However, dense point clouds face challenges of high memory consumption and reduced maintainability for long-term operations. In this study, we introduce SLIM, a scalable and lightweight mapping system for long-term LiDAR mapping in urban environments. The system begins by parameterizing structural point clouds into lines and planes. These lightweight and structural representations meet the requirements of map merging, pose graph optimization, and bundle adjustment, ensuring incremental management and local consistency. For long-term operations, a map-centric nonlinear factor recovery method is designed to sparsify poses while preserving mapping accuracy. We validate the SLIM system with multi-session real-world LiDAR data from classical LiDAR mapping datasets, including KITTI, NCLT, HeLiPR and M2DGR. The experiments demonstrate its capabilities in mapping accuracy, lightweightness, and scalability. Map re-use is also verified through map-based robot localization. Finally, with multi-session LiDAR data, the SLIM system provides a globally consistent map with low memory consumption ($\sim$130 KB/km on KITTI).
\end{abstract}


\begin{IEEEkeywords} 
Mobile Robot, LiDAR, Long-term SLAM, Nonlinear Factor Recovery.
\end{IEEEkeywords}


\section*{Supplementary Materials}

\indent A video is submitted as a multimedia attachment. A public version can be accessed \href{https://youtu.be/8HQnYMf_BWI}{online}. Code is released at \href{https://github.com/HKUST-Aerial-Robotics/SLIM}{https://github.com/HKUST-Aerial-Robotics/SLIM}.

\input{sections/introduction}

\input{sections/related_work}

\input{sections/system_overview}

\input{sections/mapping}

\input{sections/nfr}

\input{sections/experiments}

\input{sections/limitations}

\input{sections/conclusion}

\input{sections/appendix}

\bibliographystyle{IEEEtran}
\bibliography{root}

\vfill

\end{document}

%% file: sections/introduction.tex
\section{Introduction}
\label{sec:introduction}
\IEEEPARstart{M}{apping} is a fundamental capability for mobile robots navigating in various environments. Modern techniques, such as simultaneous localization and mapping (SLAM), provide metric maps that facilitate subsequent tasks like online map-based localization and planning. Over the past decade, light detection and ranging (LiDAR) has emerged as the gold standard for robotic mapping. Significant advancements in LiDAR-based mapping have been validated in diverse environments, such as underground areas \cite{palieri2020locus} and forests \cite{babin2021large}.

The urban environment is the primary setting of our daily lives. Mapping in urban environments has a rich history, dating back to the development of autonomous vehicles during the DARPA Urban Challenge~\cite{buehler2009darpa}. With advancements in LiDAR SLAM, mobile robots are now well-developed to handle mapping tasks in well-structured urban areas. However, deploying existing systems on mobile robots for long-term use presents several challenges in urban environments, stated as follows
\begin{enumerate}
\item LiDAR sensors generate 3D point clouds directly. Conventional mapping systems have achieved accurate robot pose estimation and LiDAR mapping using dense representations like points \cite{xu2022fast}, voxels \cite{hornung2013octomap}, and meshes~\cite{lin2023immesh}. While these methods enable precise environmental modeling, they are not memory-efficient for practical applications, particularly on resource-constrained platforms operating in large-scale urban environments.
\item High-level or implicit representations, such as objects or neural functions, offer memory efficiency but may not ensure sufficient accuracy for precise localization in urban environments. Reusing the generated map is crucial for robotic applications, making the integration of low-level geometric information essential in LiDAR mapping systems.
\item A limitation of many existing LiDAR mapping systems is their design for single-session tests, lacking consideration for long-term deployment. Scalability becomes crucial for multi-session LiDAR mapping systems in two aspects: globally merging multiple map sessions and maintaining the consistency and size of local maps as the number of sessions increases.
\end{enumerate}

Generally, urban environments are characterized by structured and manually deployed settings, such as poles and traffic signs, which provide strong prior knowledge for long-term mapping. Similar structures are found in indoor environments. Kimera \cite{rosinol2021kimera} by Rosinol \textit{et al.} modeled the semantic scene graph in such settings. Inspired by these studies, we designed a \textit{scalable} and \textit{lightweight} LiDAR mapping (SLIM) system that leverages the structuralism of urban environments. Figure~\ref{fig:teaser_image} presents the maps generated from SLIM, compared to widely used point cloud maps. Specifically, the key contributions span from the front-end representation to the back-end smoothing, as follows: 
\begin{enumerate}
\item We propose to parameterize lines and planes as memory-efficient representations, which encode geometric information and are also suitable for map merging.
\item Subsequently, pose graph optimization (PGO) and bundle adjustment (BA) are designed to refine the LiDAR mapping in a coarse-to-fine manner.
\item We introduce a map-centric marginalization scheme to manage map size as sessions increase, ensuring scalability for long-term maintenance. The key contribution is an efficient nonlinear factor recovery (NFR) method based on parameterized landmarks.
\item We validate the SLIM system on four different multi-session datasets in terms of accuracy, lightweightness, and scalability.
\end{enumerate}

The proposed SLIM system is an extended version of our previous work in \cite{yu2023multi}. This enhanced version introduces improvements across three key aspects. Firstly, it no longer relies on semantic information for line and plane extraction, focusing solely on geometric characteristics. We believe that additional semantic information from visual perception can enhance the mapping performance in practice. Secondly, the map merge module has been redesigned to improve robustness and efficiency for urban mapping. Lastly, we introduce a map-centric nonlinear factor recovery method to improve the system's scalability. In summary, these improvements make the SLIM system more applicable for long-term urban mapping.

\input{figure-caption/teaser_image}

The rest of our paper is organized as follows: Section~\ref{sec:relatedwork} presents related work on LiDAR mapping approaches. Section~\ref{sec:overview} provides an overview of the designed SLIM system. Section~\ref{sec:mapping} details the representations, map merge, and optimization within the SLIM system. Section~\ref{sec:nfr} introduces our proposed map-centric marginalization for long-term operation. We validate the proposed system in real-world scenarios in Section~\ref{sec:exp}. Discussions and limitations are presented in Section~\ref{sec:limitations} for in-depth analysis. The findings and future studies are summarized in Section~\ref{sec:conclusion}.

%% file: figure-caption/teaser_image.tex

\begin{figure}[t]
  \centering
    \includegraphics[width=0.98\linewidth]{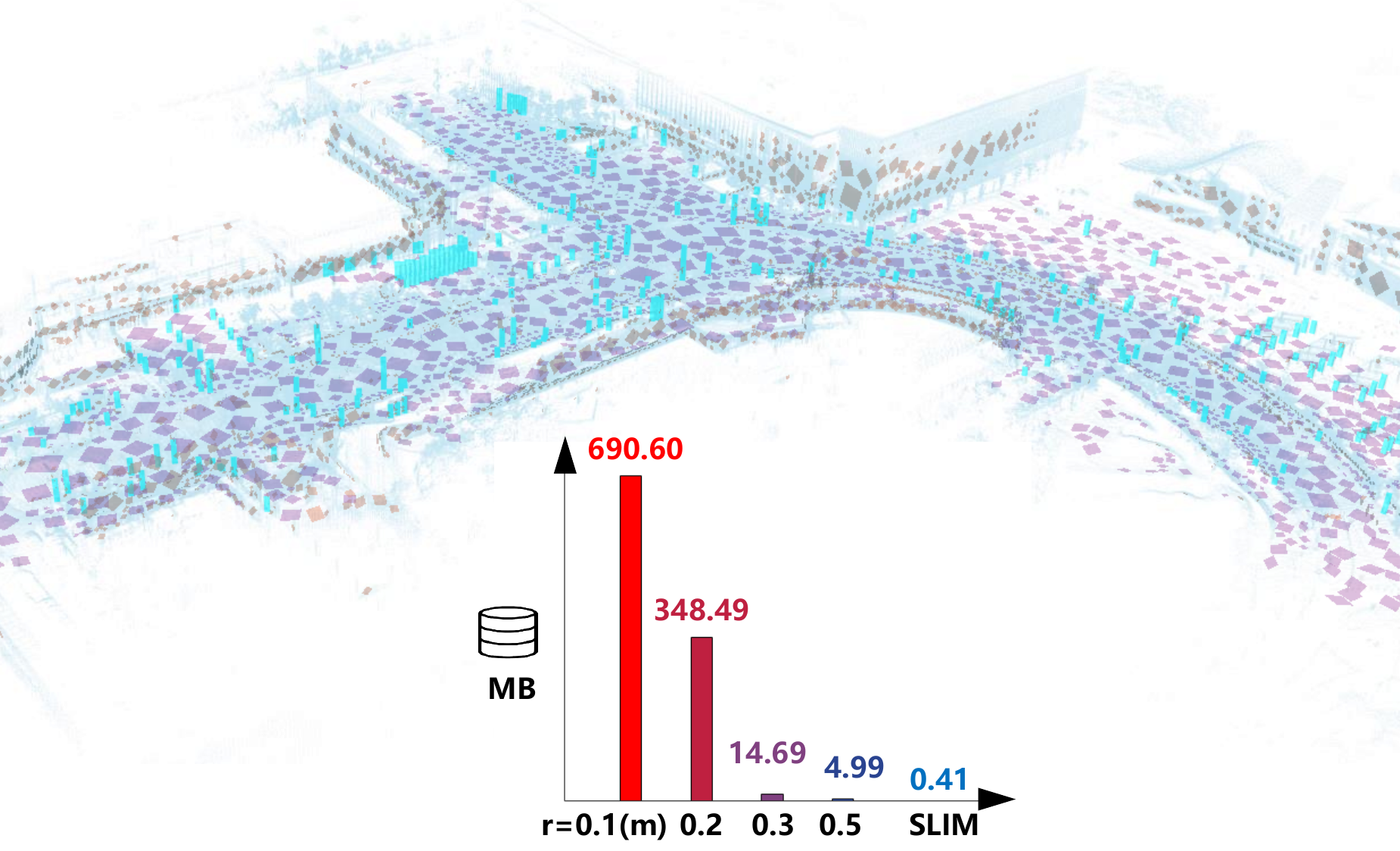}
  \caption{Demonstration of SLIM using the HeLiPR dataset~\cite{jung2023helipr}. Best viewed zoomed in and in color. SLIM provides parameterized maps with lines (colored in cyan) and planes (colored in magenta). For comparison, we also present point clouds (in light blue), which are widely used in conventional LiDAR mapping systems. More mapping results are presented in Figure~\ref{fig:mapping_visualization}. Furthermore, we compare the memory consumption of SLIM-generated maps with point cloud maps of varying densities (downsampled with radius $r$). In this HeLiPR session, SLIM maps are inherently more lightweight than LiDAR point cloud maps, with memory usage significantly reduced from 690.60 MB (point clouds) to just 0.41 MB (SLIM).}
  \label{fig:teaser_image}
\end{figure}

%% file: sections/related_work.tex
\section{Related Work}
\label{sec:relatedwork}

This section first focuses on multi-session LiDAR SLAM, a prominent research topic over the past five years. Subsequently, we discuss the representations of LiDAR SLAM and long-term mapping techniques.

\subsection{Multi-session LiDAR SLAM}

Research on LiDAR SLAM has achieved significant success. Representative works include LOAM \cite{zhang2014loam}, Fast-LIO2~\cite{xu2022fast}, DLO \cite{chen2022direct}, and BALM2 \cite{liu2023efficient}. These can be categorized into two types: the majority \cite{xu2022fast, chen2022direct, zhang2014loam, pan2021mulls, vizzo2023kiss} adopt a scan-to-map scheme with a fixed map, demonstrating high efficiency and accuracy. The second branch is BA-based methods, which parameterize map points as planes~\cite{kaess2015simultaneous, liu2023efficient, zhou2023efficient, ferrer2019eigen}, occupancy fields \cite{zhao2022occupancy}, and implicit maps \cite{pan2024pin}. BA-based methods optimize the map along with the pose to achieve higher accuracy compared to scan-to-map approaches. As LiDAR SLAM has developed, researchers have shifted focus towards multi-session LiDAR SLAM. This involves assembling individual maps and poses from multiple sessions to construct a more complete and accurate globally consistent map, under either a distributed or centralized scheme.

The general pipeline for assembling multi-session maps in different coordinates involves two key steps: loop closure detection and false loop pruning \cite{yin2024survey}. These steps aim to establish correspondences and eliminate incorrect ones, respectively. Several methods employ different approaches for loop closure detection. Maplab \cite{schneider2018maplab} and DiSCo-SLAM~\cite{huang2021disco} utilize the Scan Context descriptor \cite{kim2021scan} to detect loops between maps. However, handcrafted methods may struggle with significant viewpoint changes and have limited descriptor capabilities. In contrast, AutoMerge \cite{yin2023automerge} uses a more powerful deep learning-based method, an improved version of FusionVLAD \cite{yin2021fusionvlad}, to detect loops in large-scale environments. To prune false loops, various heuristics are introduced to reject potential false positives, including quality assessment of point cloud registration \cite{vidanapathirana2023spectral}, loop closure prioritization \cite{denniston2022loop}, and ensuring sequential consistency \cite{liu2019seqlpd}. While heuristic methods provide a rapid means of rejecting obvious false loops, they can still suffer from appearance ambiguities. Therefore, most multi-session LiDAR SLAM approaches \cite{huang2021disco, lajoie2020door, tian2022kimera, ebadi2020lamp} employ pairwise consistent measurement (PCM) set maximization \cite{mangelson2018pairwise} to prune outliers. This involves finding the maximum geometrically consistent loop set, thereby decreasing false loop associations with similar appearances.

\indent To ensure the global consistency of the merged map, PGO is employed to align the trajectories from individual sessions and to reduce the accumulated drift caused by LiDAR odometry. Latif \textit{et al.} \cite{latif2013robust} introduce a novel algorithm, Realizing, Reversing, and Recovering (RRR), to identify and reject loops that violate odometry constraints. To improve efficiency, LTA-OM \cite{yuan2023lta} eliminates the need for re-estimating graph parameters by evaluating the residual outcomes of key point factors and determining whether to recover the backup graph based on these residuals. However, these methods, which involve the incremental addition and rejection of loops, face efficiency limitations due to the requirement of multiple optimization rounds. To address this issue, contemporary multi-session SLAM systems, such as Kimera-Multi \cite{tian2022kimera} and LAMP 2.0 \cite{chang2022lamp}, incorporate all loop factors into the M-Estimator optimization to reject outliers. Subsequently, Graduated Non-Convexity (GNC) \cite{yang2020graduated} is employed to achieve robust PGO. In contrast to the aforementioned approaches that focus on global consistency using PGO, HBA~\cite{10024300} employs LiDAR BA to enhance map consistency directly and introduces a hierarchical structure to reduce computation cost.

\subsection{Representations for LiDAR SLAM}

Map representation is a front-end and crucial component of LiDAR SLAM. In this subsection, we categorize map representations into two types: explicit and implicit. Modeling the world with these representations is closely coupled with specific robotic tasks. For multi-session LiDAR mapping in this study, it is essential to consider both map memory consumption and localizability.

\subsubsection{Explicit representation} The explicit representations can be further categorized into dense spatial and sparse landmark-based representations. Dense representations typically include dense points \cite{tang2023thma}, voxels \cite{wan2018robust}, mesh \cite{lin2023immesh}, surfels \cite{chen2019suma++}, and TSDF \cite{kuhner2020large}. These aim to construct comprehensive and detailed descriptions of occupied spaces or surfaces. However, dense representations could be a huge burden for map transmission and storage in multi-session SLAM, often containing redundant information, such as complete occupancy details, which are unnecessary for urban navigation. Conversely, sparse representations, such as lines~\cite{schaefer2019long, li2021robust, pauls2021automatic, wen2022roadside}, planes~\cite{liu2021balm, liu2023efficient, ferrer2019eigen, zhou2022mathcal}, quadrics~\cite{xia2023quadric}, and segments~\cite{dube2020segmap}, could be better choices to describe large-scale urban environments. Among these, line and plane representations leverage the inherent characteristics of urban scenes, balancing high localization accuracy with minimal memory requirements.

\subsubsection{Implicit representation} Implicit representations in radiance~\cite{mildenhall2021nerf}, distance~\cite{park2019deepsdf}, and occupancy~\cite{mescheder2019occupancy} fields have significantly influenced LiDAR SLAM research. SHINE-Mapping~\cite{zhong2023shine} presents an incremental implicit mapping system incorporating online learning of sparse octree node features. LONER~\cite{isaacson2023loner} adopts a hierarchical feature grid encoding inspired by Ins-NGP, building a real-time neural implicit LiDAR SLAM system enhanced with LiDAR odometry. NeRF-LOAM~\cite{deng2023nerf} introduces a dynamic voxel embedding generation approach designed for large-scale scenario. These implicit neural LiDAR SLAM methods exhibit impressive mapping accuracy. However, they either necessitate mesh map reconstruction for localization or depend directly on the implicit map, constraining their operational frequency for real-time use and long-term scalable mapping.



\subsection{Long-term Mapping}

Long-term mapping presents challenges for visual-based frameworks due to illumination changes~\cite{toft2020long,wenzel20214seasons}. Unlike visual sensing, LiDAR sensing remains consistent and illumination-invariant across day-night variations, making it well-suited for long-term mapping. At the front end, managing newly generated maps from robots has been extensively studied. Typically, map merging involves place recognition and point cloud registration techniques~\cite{yin2024survey}. For long-term managment, compressing point cloud size can reduce computational demands for online applications, utilizing observation-based~\cite{yin20203d} and geometric-based sampling~\cite{labussiere2020geometry}.

Most works in the community focus on back-end processing for long-term mapping, specifically on the SLAM problem. Early filter-based methods, such as sparse extended information filters, demonstrated that SLAM could be solved in constant time by exploiting sparse structures~\cite{thrun2004simultaneous,vial2011conservative}. Modern SLAM is typically formulated as an optimization problem~\cite{cadena2016past} using pose graphs (only poses) or factor graphs (poses and landmarks)~\cite{grisetti2010tutorial,dellaert2012factor}. In pose graph-based SLAM, sensing information on poses can compact the graph structure\cite{ila2009information,kretzschmar2012information}. Kretzschmar and Stachniss~\cite{kretzschmar2012information} design an information-aided long-term 2D LiDAR mapping scheme where laser measurements with minimal information are discarded, and pose nodes are marginalized accordingly. The researchers use a Chow-Liu tree to maintain the marginalized pose graph. The main challenge with these approaches is modeling information (uncertainty), especially for dense 3D LiDAR scans. GLC~\cite{carlevaris2014generic} is developed to remove nodes from SLAM graphs, bounding computational complexity for multi-session mapping in the same area. Recent work by Doherty \textit{et al.}~\cite{doherty2022spectral} proposes selecting measurements by maximizing the algebraic connectivity of pose graphs. Similar measurement selection techniques have been designed in previous studies~\cite{khosoussi2019reliable,kurz2021geometry} for long-term graph pruning.

In the NFR by Mazuran \textit{et al.}\cite{mazuran2014nonlinear,mazuran2016nonlinear}, graph sparsification is formulated as a Kullback–Leibler divergence minimization problem, with a closed-form solution available. The mean and covariance of maintained variables are re-estimated, i.e., recovered. The prior condition for NFR is the existence of a node removal method. This minimization can also be approached in a descent form, as shown in~\cite{vallve2018graph}. Jiang and Shen~\cite{jiang2024two} propose a new NFR framework for long-term visual mapping, focusing on landmark-based factors. Relevant back-end techniques have been applied to several SLAM systems, such as cooperative SLAM for marine robots~\cite{paull2015communication} and visual-inertial odometry~\cite{usenko2019visual}. Inspired by these studies, we adopt a map-centric NFR into our SLIM system for long-term map maintenance. Considering the sparsity of urban environments, the closed-form solution is approximated using sparse matrix operations, as detailed in Section~\ref{sec:nfr}.



%% file: sections/system_overview.tex
\section{System Overview}
\label{sec:overview}

\begin{figure*}[t]
    \centering
    \includegraphics[width=0.95\linewidth]{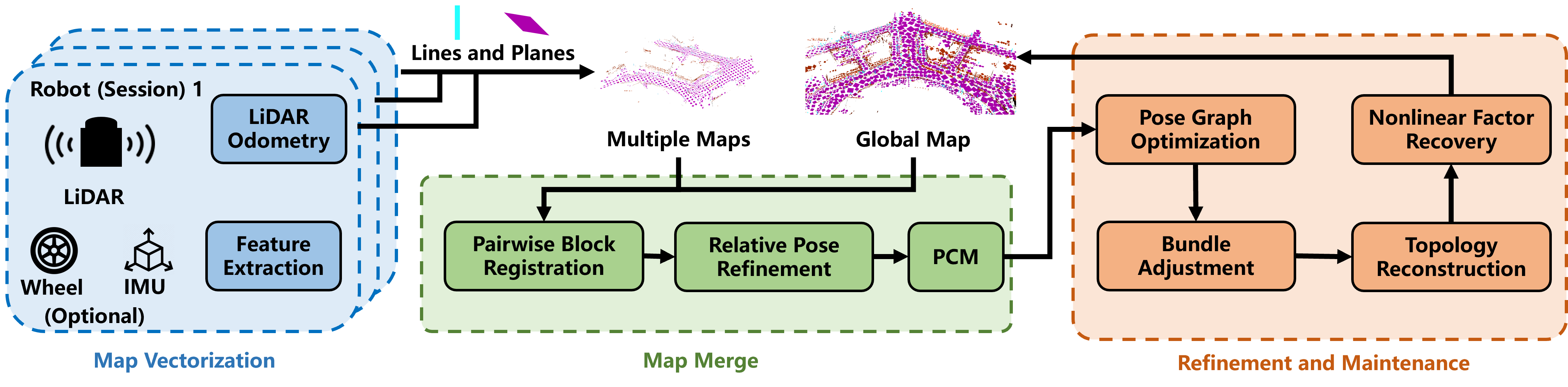}
    \caption{System Overview. The front-end map vectorization module extracts features and parameterizes planes to lanes. Then, the SLIM system generates one global lightweight map via map merge. The map refinement modules, including PGO and BA, smooth the mapping results only using lines and planes. The map-centric marginalization can bound the computational and storage requirements with increasing map sessions. Overall, SLIM is a centralized server that can achieve map vectorization, map merge and refinement for long-term LiDAR mapping.}
    \label{fig:system}
\end{figure*}

The system pipeline is illustrated in Figure~\ref{fig:system}. Given sequential input LiDAR scans, SLIM first converts raw LiDAR point clouds into line and plane representations based on geometric properties. This vectorization pipeline is detailed in Section~\ref{sec:representations}. These parameterized representations are compact and manageable for subsequent tasks. Lines and planes are accumulated using LiDAR odometry or with additional onboard odometry, resulting in lightweight landmarks in maps. To merge multi-session maps, SLIM utilizes a global registration approach without place recognition, detailed in Section~\ref{sec:merge}. Though the maps are unified in the same coordinate, the drift from odometry still exists. In Section~\ref{sec:refinement}, SLIM achieves map refinement (smoothing) through a coarse-to-fine approach: PGO and BA, designed based on the parameterized lines and planes. With increased merged map sessions, a map-centric nonlinear factor recovery is implemented to keep the number of poses manageable, as presented in Section~\ref{sec:nfr}. All these operations mentioned above make the mapping pipeline scalable with increasing multi-session LiDAR maps. We also summarize the pipeline in Algorithm~\ref{alg:system}. It is worth noting that the map landmarks consist solely of parameterized lines and planes after map vectorization, without storing any dense point clouds in the map. 



\begin{algorithm}[t]
    \caption{\textbf{SLIM Pipeline}}
    \label{alg:system}
    \setlength{\baselineskip}{1.2\baselineskip}
    \SetKwInOut{Input}{Input}
    \SetKwInOut{Output}{Output}
    
    \Input{Base Map $\mathcal{M}_{b,{k-1}}$ at $k-1$, New submap $\mathcal{M}_s$}
    \Output{New Base Map $\mathcal{M}_{b,k}$ at $k$}
    $\left\{ \mathcal{G}_{b_p} \right\}, \left\{ \mathcal{G}_{s_q} \right\} = \texttt{partition}(\mathcal{M}_{b,k-1}, \mathcal{M}_s)$ \ref{sec:mapmergestrategy} \\
    $\left\{\mathbf{T}^{b_i}_{s_i}\right\} = \texttt{registerBlock}(\left\{ \mathcal{G}_{b_p} \right\}, \left\{ \mathcal{G}_{s_q} \right\})$ \ref{sec:graphregistration}\\
    $\left\{\mathbf{T}^{b_i}_{s_i}\right\} = \texttt{filterLoop}(\left\{\mathbf{T}^{b_i}_{s_i}\right\})$ \ref{sec:pcm}\\
    $\mathcal{\bar M}_{b,k} = \texttt{PGO}(\mathcal{M}_{b,k-1}, \mathcal{M}_{s}, \left\{\mathbf{T}^{b_i}_{s_i}\right\})$ \ref{sec:PGO} \\
    $\mathcal{\bar M}_{b,k} = \texttt{BA}(\mathcal{\bar M}_{{b,k}})$ \ref{sec:localBA} \\
    $\mathcal{M}_{b,k} = \texttt{Marginalization}(\mathcal{\bar M}_{{b,k}})$ \ref{sec:nfr} \\
\end{algorithm}

%% file: sections/mapping.tex
\section{Map Representation, Merge and Refinement}
\label{sec:mapping}

We first present the front-end map vectorization in Section~\ref{sec:representations}. Subsequently, the map merge function is introduced in Section~\ref{sec:merge} to unify local maps within the global frame through place recognition and global registration. In Section~\ref{sec:refinement}, map refinement modules improve mapping accuracy using our proposed PGO and BA approaches.

\subsection{Map Vectorization: From Point Clouds to Lines and Planes}
\label{sec:representations}

\subsubsection{Pre-processing and Feature Points Selection} 
\label{sec:preprocessing}

Given sequential LiDAR scans from robots, we apply existing LiDAR odometry methods, such as LOAM~\cite{zhang2014loam} and KISS-ICP~\cite{vizzo2023kiss}, to obtain local maps and sequential robot poses. Sensor fusion approaches~\cite{xu2022fast} could also be an alternative. These methods provide dense point clouds and robot poses with drifts, serving as input for the SLIM system. Typically, robots or vehicles operate in different parts of urban environments, corresponding to different local frames. The SLIM system is designed to provide a consistent and lightweight map in a global frame.



The most common objects in urban environments are buildings, roads, and roadside poles, which can be categorized into two types of features: \textit{lines} and \textit{planes}. For line feature extraction, we initially divide points into their respective scan lines to restore the original scanning pattern. It is important that points on the same scan line are arranged in the order of scanned timestamp. Subsequently, we calculate the bilateral distance difference for a point on a scan line. Based on the distance difference values on both sides, we label points as negative gradient points, positive gradient points, and points with gradients on both sides with a threshold. We then identify pairs of neighboring negative and positive gradient points that are sufficiently close in distance. Extracting the points between them, as well as those with gradients on both sides, allows us to obtain a candidate point set. Finally, we employ clustering and line fitting methods to refine this candidate point set, finally obtaining multiple \textit{line segments}. Planar points are selected based on the local structure of points. First, TRAVEL~\cite{oh2022travel} is used to identify ground and non-ground points. Hash octrees are then constructed separately for these two different types of points. Classical principal component analysis (PCA) is applied to identify planar points and convert them to \textit{plane segments}. These line and plane segments are significantly fewer than the raw LiDAR points while preserving the original geometric information.


We define a \textit{line observation} as $\mathbf{f}_{\mathcal{L}}:=\langle \mathbf{\hat p}_a,\mathbf{\hat p}_b, \sqrt{\mathbf{\Lambda}_{\mathcal{L}}}, N \rangle$, and a \textit{plane observation} as $\mathbf{f}_{\mathcal{S}}:=\langle \mathbf{\hat p}_a,\mathbf{\hat p}_b, \mathbf{\hat p}_c, \sqrt{\mathbf{\Lambda}_{\mathcal{S}}}, N \rangle$. Two observation points are used to build residuals for line landmarks, while three are used for plane landmarks. To ensure that these selected sparse observation points effectively represent the local point cloud structure, their selection is based on the eigenvalue decomposition results of a local PCA. More specifically, we assume that all extracted LiDAR points of a single observation is $\left\{\mathbf{p}_i\right\}$ and their  centroid is $\mathbf{\bar p}$. The computed eigenvalues are $\lambda_1 <\lambda_2< \lambda_3$ and corresponding eigenvectors are $\mathbf{v}_1, \mathbf{v}_2, \mathbf{v}_3$. The two line observation points are defined as
\begin{equation}
\begin{aligned}
    \mathbf{\hat p}_a &= \mathbf{\bar p} + \sqrt{2\lambda_3}\mathbf{v}_3  \\
    \mathbf{\hat p}_b &= \mathbf{\bar p} - \sqrt{2\lambda_3}\mathbf{v}_3 
\end{aligned}
\label{eq:LineObDefinition}
\end{equation}
\indent Similarly, the three plane observation points are defined as
\begin{equation}
\begin{aligned}
    \mathbf{\hat p}_a &= \mathbf{\bar p} + \sqrt{2\lambda_2}\mathbf{v}_2\\
    \mathbf{\hat p}_b &= \mathbf{\bar p} - \sqrt{\frac{\lambda_2}{2}}\mathbf{v}_2 + \sqrt{\frac{\lambda_1}{2}}\mathbf{v}_1 \\
    \mathbf{\hat p}_c &= \mathbf{\bar p} - \sqrt{\frac{\lambda_2}{2}}\mathbf{v}_2 - \sqrt{\frac{\lambda_1}{2}}\mathbf{v}_1
\end{aligned}
\label{eq:LineObDefinition}
\end{equation}
\indent It is worth noting that the degrees-of-freedom (DoF) for the line residuals of two observation points are 4, which is consistent with the DoF of line landmarks. Similarly, the DoF for the plane residuals of three observation points are 3, which is consistent with the DoF of plane landmarks. This design is optimal for our LiDAR bundle adjustment and scalable mapping, as further elaborated in Sections \ref{sec:mapping} and \ref{sec:nfr}.

\indent We also introduce information matrices to describe uncertainties of extracted lines and planes. We define $\sqrt{\mathbf{\Lambda}} = \mathbf{I}\cdot(\sqrt{N/m}/\sigma)$ as the square root of the information matrix, where $(\sqrt{\mathbf{\Lambda}})^T\sqrt{\mathbf{\Lambda}}=\mathbf{\Lambda}$ and $m$ is the number of observation points for a landmark. The term $N$ represents the number of raw points in line or plane segments. The variance $\sigma$ varies across different semantic categories. Specifically, $\sigma$ is determined by the flatness of point cloud. We set $\sigma_r=0.1, \sigma_b=0.2, \sigma_p=0.3$ for roads, buildings, and poles in the experimental validation.






\subsubsection{Vectorized Mapping}
\label{sec:mapvectorization}


Parameterizing the \textit{observations} into \textit{landmarks} is crucial for long-term maintenance. To reduce the complexity, we sample dense robot poses from odometry as \textit{keyframes}. Regarding the landmarks, we define a \textit{line landmark} as $\mathbf{l}_{\mathcal{L}} := \langle l_l, \mathbf{c}_{\mathcal{L}}, \mathbf{n}_{\mathcal{L}}, \mathbf{p}_{\mathcal{L}}, \left\{ \mathbf{f}_{{\mathcal{L}}_i} \right\} \rangle$ and a \textit{plane landmark} as $\mathbf{l}_{\mathcal{S}} := \langle l_s, \mathbf{c}_{\mathcal{S}}, \mathbf{n}_{\mathcal{S}}, \mathbf{p}_{\mathcal{S}}, \left\{ \mathbf{f}_{{\mathcal{S}}_i} \right\} \rangle$, which include a label, centroid, normal, minimum parameter block, and their \textit{observations} across different \textit{keyframes}. Specifically, the normal is a directional vector for a line and a normal vector for a plane. The \textit{observations} encodes that the landmark is associated with a specific \textit{keyframes}, and this association can be obtained through centroid-based nearest neighbor searching when applying LiDAR odometry. Essentially, we have constructed a factor graph structure similar to \textit{visual SLAM}.


The next step is to parameterize the landmarks to enable optimization modules for map refinement. Meanwhile, a minimum parameter block is needed to reduce the problem scale as mentioned. The point-normal form, i.e., the normal vectors $\mathbf{n}$ and random points $\mathbf{c}$, is over-parameterized for lines and planes and unsuitable for optimization problems. To address this, we propose representing an infinite line as $\mathbf{p}_{\mathcal{L}} := \langle \alpha, \beta, x, y \rangle \in \mathbb{R}^4$, where $\alpha$ and $\beta$ represent the direction, and $x$ and $y$ represent the offset translation on the $xOy$ plane. Similarly, an infinite plane is formulated as $\mathbf{p}_{\mathcal{S}} := \langle \alpha, \beta, d \rangle \in \mathbb{R}^3$, where $\alpha$ and $\beta$ represent the direction, and $d$ represents the offset translation on the $z$-axis, thus formulating the desired minimum parameter blocks.


\begin{figure}[t]
	\centering
	\includegraphics[width=0.95\linewidth]{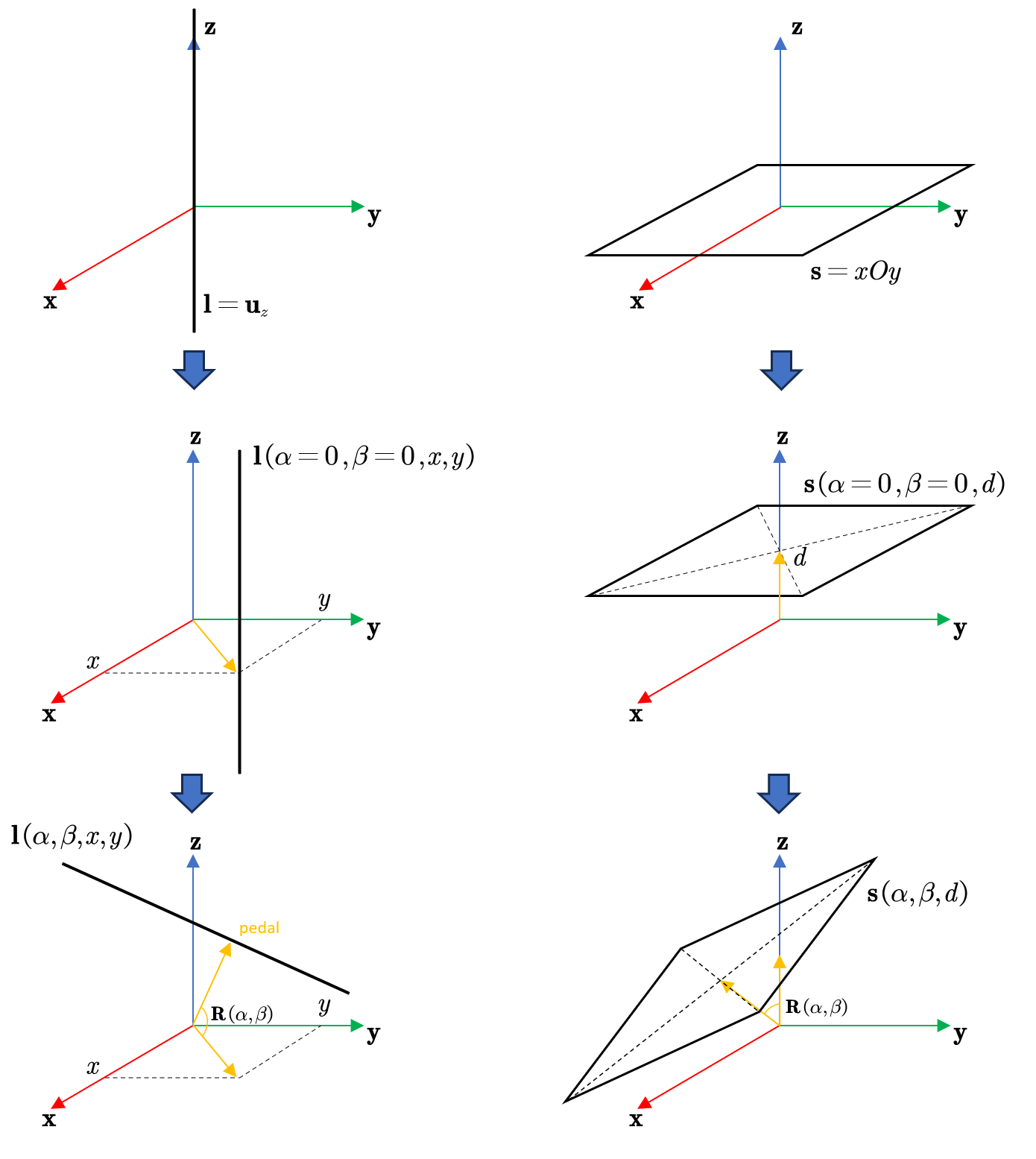}
	\caption{Formulation of line and plane landmarks. On the left side, the line $\mathbf{l}(x, y)$ can be obtained through a translation $(x, y)$, followed by the derivation of $\mathbf{l}(\mathbf{R}(\alpha, \beta), x, y)$ using a 2-DoF rotation $\mathbf{R}(\alpha, \beta)$. Similarly, on the right side, the plane $\mathbf{s}(d)$ is formulated by applying a translation $d$, and subsequently transformed into $\mathbf{s}(\mathbf{R}(\alpha, \beta), d)$ via rotation. A corresponding transformation always exists for any line or plane to formulate landmarks.}

	\label{figure:landmark_formulation}
\end{figure}

Figure \ref{figure:landmark_formulation} illustrates how we obtain arbitrary lines and planes using our minimal parameters. Theoretically, all infinite lines and infinite planes could be transformed from the original line $\mathbf{u}_z$ and the original plane $xOy$ in two steps, where $\mathbf{u}_x,\mathbf{u}_y,\mathbf{u}_z$ are unit vectors along $x$-axis, $y$-axis and $z$-axis. For line landmarks, we first apply an offset $(x, y)$ to the original line $\mathbf{u}_z$ to obtain the translated line $\mathbf{l}(x, y)$. Then, we rotate the nearest point $(x, y)$ using one 2 DoF rotation matrix $\mathbf{R}(\alpha, \beta)$ to obtain the final line $\mathbf{l}(\alpha, \beta, x, y)$. The 2 DoF rotation matrix $\mathbf{R}(\alpha, \beta)$ is defined as follows:
\begin{equation}
    \mathbf{R}\left( \alpha ,\beta \right) =\left[ \begin{matrix}
    	\cos \beta&		0&		-\sin \beta\\
	\sin \alpha \sin \beta&		\cos \alpha&		\sin \alpha \cos \beta\\
	\cos \alpha \sin \beta&		-\sin \alpha&		\cos \alpha \cos \beta\\
    \end{matrix} \right] 
    \label{eq:Rotation2dof}    
\end{equation}
\par For plane landmarks, the original plane $xOy$ is translated along the $z$ axis and rotated using a 2 DoF rotation matrix, resulting in the final plane $\mathbf{s}(\alpha, \beta, d)$. 

The minimum parameterization of lines and planes necessitates residuals related to the pose and landmark. Though point-to-line and point-to-plane residuals are available in existing studies, they can not be applied directly on the parameterized forms of this study. Hence, it is essential to establish the mapping relations between our designed representations and the conventional point-normal form. The mappings are stated as follows:
\begin{equation} 
    \begin{bmatrix}
    \mathbf{n}_{\mathcal{L}} \\ \mathbf{c}_{\mathcal{L}}
    \end{bmatrix}
    =
    \begin{bmatrix}
        \mathbf{R}(\alpha, \beta)\mathbf{u}_z \\
        \mathbf{R}(\alpha, \beta)(\mathbf{u}_x x + \mathbf{u}_y y)
    \end{bmatrix}
    \label{eq:LineConvert}
\end{equation}
\begin{equation} 
    \begin{bmatrix}
    \mathbf{n}_{\mathcal{S}} \\ d_{\mathcal{S}}
    \end{bmatrix}
    =
    \begin{bmatrix}
        \mathbf{R}(\alpha, \beta)\mathbf{u}_z \\
        d
    \end{bmatrix}
    \label{eq:SurfConvert}
\end{equation}

Detailed residual formulations, utilizing the proposed line and plane representations, are presented in Appendix~\ref{sec:appendix}. The optimization process (map refinement) is introduced in Section~\ref{sec:refinement} and further illustrated in Appendix~\ref{sec:appendix}.

\subsection{Global Map Merge}
\label{sec:merge}

\subsubsection{Map Merge Strategy}
\label{sec:mapmergestrategy}

To this end, local maps are parameterized with lines and planes from multi-session data. We refer to these local maps as \textit{submaps} $\left\{\mathcal{M}_{s_i}\right\}$ in the remainder of this paper. These submaps are located in different places and under different coordinate systems. Consequently, the global map merging process entails aligning the submaps to a common coordinate system, merging and eliminating redundant landmarks, and generating a globally consistent map that will serve as the base map for subsequent merges. We denote the merged global map as the base map $\mathcal{M}_b$. At the beginning of the merging process, we can select any submap to serve as the initial base map. 

Typical global map merge methods first transform full laser scans into handcrafted or learning-based global descriptors for place retrieval. Then, point cloud registration is indispensable for relative transformation estimation, as introduced in Section~\ref{sec:relatedwork}. Existing approaches often require extensive raw information for place recognition, such as the original point cloud or embedding feature maps, leading to significant data demands for learning and transmission. Additionally, these methods mostly focus on scan-to-scan place retrieval rather than utilizing multi-frame information to encode a place. Single scan-based methods are more sensitive to displacement and lack robustness~\cite{yin2024survey}.



In this study, we propose registering submaps directly for global map merging \textit{without} place recognition. Specifically, we adopt our previous work, G3Reg~\cite{qiao2023g3reg}, for fast and robust registration. G3Reg utilizes graph theory, specifically the Maximum Clique (MC) algorithm, to prune correspondence outliers, with the landmark correspondences acting as graph nodes. However, each submap contains numerous landmarks (as detailed in Section~\ref{sec:representations}), and solving the MC problem for such a large graph is computationally infeasible, resulting in an extremely long processing time.


To address this issue, we design a two-step method to reduce the problem size. First, we cluster the plane landmarks that lie on the same infinite plane into a single plane, particularly for those on large buildings and road surfaces. The line landmarks are maintained without clustering. Second, we partition the base map into blocks $\left\{\mathcal{G}{b_p}\right\}$ and the submap into blocks $\left\{\mathcal{G}{s_q}\right\}$ along the robot trajectory, instead of directly matching the submap to the base map. Each block comprises a host keyframe and a set of landmarks around it. These two steps make the map structure more compact for finding the maximum clique by reducing the number of nodes in a block, thereby leading to more efficient block registration.


\subsubsection{Pairwise Block Global Registration}
\label{sec:graphregistration} 



Estimating correct correspondences is crucial for global block registration. G3Reg proposes Gaussian ellipsoid models on landmarks, which consist of a centroid and its uncertainty represented by a pseudo-covariance matrix, to construct Translation and Rotation Invariant Measurements (TRIMs). These TRIMs are then used to build a compatibility graph for correspondence outlier pruning. 

In this study, the basic representations are parameterized infinite lines and planes without centroid modeling in Euclidean space. Such parameterization cannot support the construction of TRIMs used in the original G3Reg~\cite{qiao2023g3reg} for outlier pruning. To address this issue, we introduce the Grassmannian metric~\cite{lusk2022graffmatch} to compute TRIMs and use the pairwise compatibility test to determine whether two correspondences are compatible. Specifically, lines and planes can be formulated as $k$-dimensional subspaces, where lines belong to one-dimensional subspaces and planes belong to two-dimensional subspaces. Their Grassmannian coordinates ${Y}$ are written as follows:
\begin{equation} 
    {Y} = \begin{bmatrix}
            A & \frac{{b}}{\sqrt{\lVert {b} \rVert^2+1}} \\
            0 & \frac{1}{\sqrt{\lVert {b} \rVert^2+1}}
        \end{bmatrix} 
         \in \mathbb{R}^{(n+1)\times(k+1)}
    \label{eq:GraffCoordinate}
\end{equation}
where $A$ and $b$ are the orthonormal basis and orthonormal displacement of the line nodes and plane nodes; $n$ represents the dimension of Euclidean space ($n=3$). Grassmannian metric $\mathrm{d}_{\mathrm{Graff}}({Y}_1, {Y}_2)$ can be defined to compute the distance of two subspaces:
\begin{equation} 
    \begin{aligned}
    b_{02}&=b_2-b_1 \\
    Y^{'}_1 &= \begin{bmatrix}
            A_1 & 0 \\
            0 & 1
        \end{bmatrix}   \\  
    Y^{'}_2 &= \begin{bmatrix}
            A_2 & b_{02}/\sqrt{\lVert b_{02} \rVert^2 + 1} \\
            0 & 1/\sqrt{\lVert b_{02} \rVert^2 + 1}
        \end{bmatrix} \\
    \left\{\sigma_i\right\} &= \mathrm{SVD}((Y^{'}_1)^TY^{'}_2)    \\
    \mathrm{d}_{\mathrm{Graff}}({Y}_1, {Y}_2) &= \
    \sum_{i}{\mathrm{arccos}^2(\sigma_i)} 
    \end{aligned}
\end{equation}



\begin{figure}[t]
  \centering
    \includegraphics[width=0.47\linewidth]{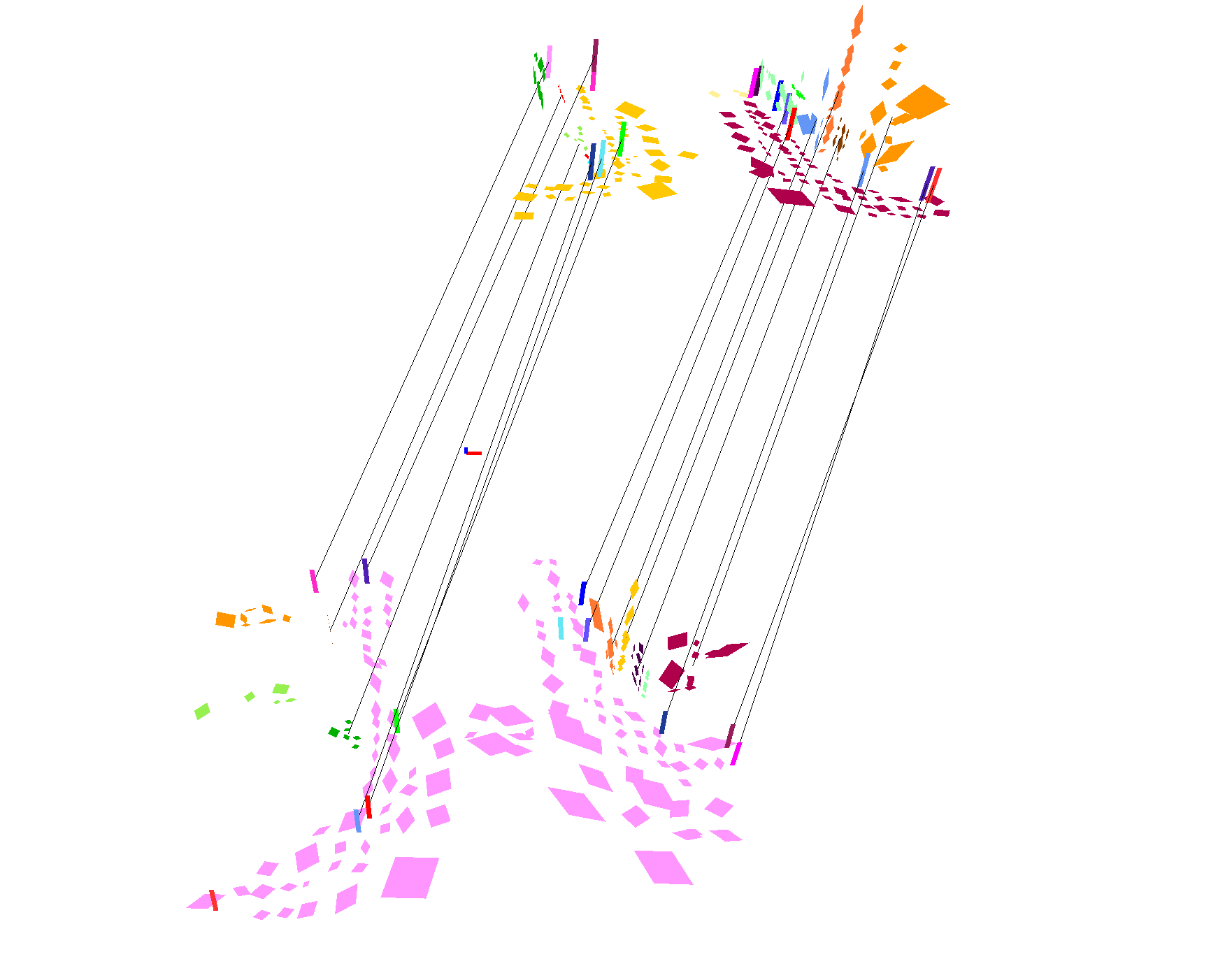}
    \includegraphics[width=0.47\linewidth]{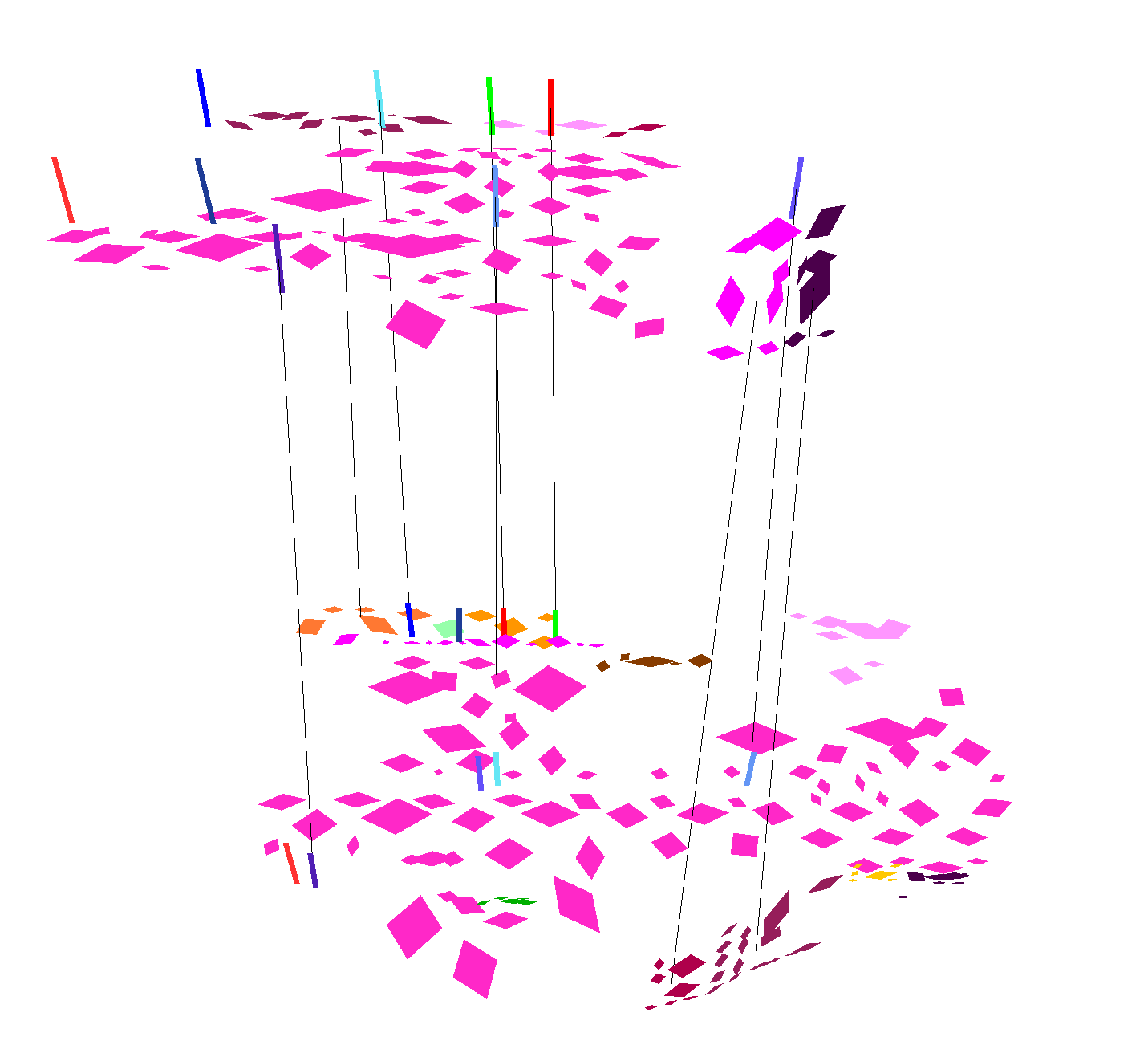}
  \caption{Two registration cases at different places, using the proposed block registration method. We reduce the number of plane landmarks by clustering, thus speeding up the registration process for map merge. The lines between the clustered landmarks represent the data associations (correspondences) between the block in different coordinates. 
  }
  \label{fig:blockreg}
\end{figure}

We denote all line and plane correspondences as $\mathcal{C}_{\mathcal{L}} = \{ (k,l) \}$ and $\mathcal{C}_{\mathcal{S}} = \{ (k,l) \}$, where $k$ and $l$ are the indices of landmarks in blocks $\mathcal{G}_{b_p}$ and $\mathcal{G}_{s_q}$ (blocks belong to the base map and the submap, respectively). The pairwise compatibility test for any two correspondences is formulated as follows:
\begin{equation}
\label{eq:comp_test}
\left| \, \text{d}_{\text{Graff}}\left( Y_{k_i},Y_{k_j} \right) - \text{d}_{\text{Graff}}\left( Y_{l_i},Y_{l_j} \right) \, \right| < \delta,
\end{equation}
where $(k_i,l_i)$ and $(k_j,l_j) \in \mathcal{C}_{\mathcal{L}} \cup \mathcal{C}_{\mathcal{S}}$. Since the Grassmannian metric is $\text{SE}(3)$-transformation invariant, we define a truncation threshold $\delta$ to determine whether both correspondences pass the compatibility test. If passed, an edge is connected between these correspondences in the compatibility graph. The maximum clique is then found using a fast exact parallel finder algorithm~\cite{rossi2013fast}, where the nodes in this clique represent the putative line and plane correspondence inlier sets $\mathcal{C}_{\mathcal{L}}^{*}$ and $\mathcal{C}_{\mathcal{S}}^{*}$. Figure~\ref{fig:blockreg} presents two cases of block registration for map merge in this study.

With the pruned correspondence results, we can establish a hybrid registration to estimate the relative poses between graphs (blocks), which is formulated as follows:
\begin{equation} 
    \begin{aligned}
        \mathop{\mathrm{min}}\limits_{\mathbf{T}^{b_i}_{s_i}}
        & \sum_{(k, l)\in \mathcal{C}_{\mathcal{L}}^{*}} \rho( \lVert (\mathbf{I} - \mathbf{n}_k\mathbf{n}^T_k)(\mathbf{T}^{b_i}_{s_i} \mathbf{c}_l - \mathbf{c}_k) \rVert_{\mathbf{\Lambda}_{\mathcal{L}}}^{2}) + \\
        & \sum_{(k, l)\in \mathcal{C}_{\mathcal{S}}^{*}} \rho( \lVert \mathbf{n}^T_k(\mathbf{T}^{b_i}_{s_i} \mathbf{c}_l - \mathbf{c}_k) \rVert_{\mathbf{\Lambda}_{\mathcal{S}}}^{2})
    \end{aligned}
    \label{eq:BlockCoarseRegistraion}
\end{equation}
where $\mathbf{c}=b$, $\mathbf{n}=A$ for lines and $\mathbf{n}\in\text{span}(A)^\perp$ for planes. The term $\mathbf{T}^{b_i}_{s_i}$ is the relative transformation between block $\mathcal{G}_{b_i}$ and block $\mathcal{G}_{s_i}$; $\rho(\cdot)$ denotes a robust kernel function, which further reduces the influence of potential correspondence outliers resulting from an inappropriate selection of the threshold $\delta$, as discussed in \cite{qiao2023g3reg}. At the start of iterations, the initial value of $\mathbf{T}^{b_i}_{s_i}$ is set to the identity transformation. 



    
    


Furthermore, an optimization-based refinement is performed based on the k-nearest neighbors (kNN) search of landmarks to refine the relative pose estimation. Note that the landmark for refinement is the vectorized ones in Section \ref{sec:mapvectorization}. The refinement is formulated as follows:
\begin{equation} 
    \begin{aligned}
        \mathop{\mathrm{min}}\limits_{\mathbf{T}^{b_i}_{s_i}}
        & \sum_{(k, l)\in \mathcal{N}_{\mathcal{L}}} \rho( \lVert (\mathbf{I} - \mathbf{n}_k\mathbf{n}^T_k)(\mathbf{T}^{b_i}_{s_i} \mathbf{p}_l - \mathbf{p}_k) \rVert_{\mathbf{\Lambda}_{\mathcal{L}}}^{2}) +\\
        & \sum_{(k, l)\in \mathcal{N}_{\mathcal{S}}} \rho( \lVert \mathbf{n}^T_k(\mathbf{T}^{b_i}_{s_i} \mathbf{p}_l - \mathbf{p}_k) \rVert_{\mathbf{\Lambda}_{\mathcal{S}}}^{2})
    \end{aligned}
    \label{eq:BlockRefineRegistraion}
\end{equation}
where $(k,l)$ is the nearest-neighbor landmark pair in different blocks with respect to the initial pose estimated from block registration in Equation~\eqref {eq:BlockCoarseRegistraion}. The terms $\mathbf{n}$ and $\mathbf{p}$ are with landmarks as demonstrated in Section \ref{sec:mapvectorization}, not the clustered landmarks for block registration. Using original landmarks could improve the accuracy of registration compared to block registration. To enhance the robustness, the optimization problem is solved iteratively with updated landmark matching, which is the pipeline in the classical iterative closest point (ICP)~\cite{besl1992method}. Finally, we will obtain a set of relative poses $\mathcal{L}^*_{\mathcal{O}}:= \left\{\mathbf{T}^{b_i}_{s_i}\right\}$ across sessions to serve as a candidate set for downstream modules.

\subsubsection{Loop Outlier Rejection}
\label{sec:pcm}

Once all blocks have been successfully registered, we can align the submaps to the global coordinate system and proceed with the subsequent map refinement steps. However, false positive loop closures still remain due to several inevitable factors, such as inaccurate landmark parameterization and outliers in correspondences. We utilize a classical approach, pairwise consistent measurement (PCM)~\cite{mangelson2018pairwise} to identify false loop candidates. Specifically, A function $C(\mathbf{T}^{b_k}_{s_k}, \mathbf{T}^{b_l}_{s_l})$ is designed to measure the consistency of block registration, based on the relative pose estimation in Section~\ref{sec:graphregistration}, formulated as follows:
\begin{equation} 
    \begin{aligned}
    \delta \mathbf{T} &= (\mathbf{T}^{b_k}_{s_k})^{-1} \cdot \mathbf{\hat T}^{s_l}_{s_k} \cdot \mathbf{T}^{b_l}_{s_l} \cdot \mathbf{\hat T}^{b_k}_{b_l} \\
        C(\mathbf{T}^{b_k}_{s_k}, \mathbf{T}^{b_l}_{s_l}) 
        &= [ \lVert \mathrm{Log}(\mathbf{R}(\delta\mathbf{T})) \rVert_2,  
        \lVert \mathbf{p}(\delta\mathbf{T}) \rVert_2]^T
    \end{aligned}
\end{equation}
where $\mathbf{\hat T}^{s_l}_{s_k}$ and $\mathbf{\hat T}^{b_k}_{b_l}$ are from odometry trajectories in different sessions. If $C(\mathbf{T}^{b_k}_{s_k}, \mathbf{T}^{b_l}_{s_l})$ are small enough, the pair of relative pose $\mathbf{T}^{b_k}_{s_k}$ and $\mathbf{T}^{b_l}_{s_l}$ are considered to be consistent. Similarly, the problem of solving the internal-consistent set of relative poses is also a maximum clique problem~\cite{rossi2013fast}. The pruned set of relative poses (transformations) is denoted as $\mathcal{L}_{\mathcal{O}}=\left\{\mathbf{T}^{b_i}_{s_i}\right\}$, and will be used for subsequent PGO.


\subsection{Map Refinement}
\label{sec:refinement}

\subsubsection{Pose Graph Optimization}
\label{sec:PGO}

For each relative pose in $\mathcal{L}_{\mathcal{O}}$, we can construct a loop residual as
\begin{equation} 
\begin{aligned}
\mathbf{r}_{\mathcal{L}_\mathcal{O}}\left(\mathbf{T}^w_{b_i}, \mathbf{T}^w_{s_i}, \mathbf{\hat T}^{b_{i}}_{s_i}\right)=
    \begin{bmatrix}
    ({\mathbf{R}^w_{b_i}})^T \left( \mathbf{p}^w_{s_i} - \mathbf{p}^w_{b_i} \right) - \mathbf{\hat p}^{b_i}_{s_i} \\
    (\mathbf{\hat R}^{b_i}_{s_i})^T (\mathbf{R}^w_{b_i})^T \mathbf{R}^w_{s_i}
    \end{bmatrix}
    \label{eq:RelPoseResidual}    
\end{aligned}
\end{equation}
where $w$ represents the world coordinate. In addition to the relative poses between blocks, the LiDAR odometry could also provide rigid transformations between consecutive keyframe poses, as illustrated in Section~\ref{sec:preprocessing}. The odometry-based residual function is defined as 
\begin{equation} 
\begin{aligned}
\mathbf{r}_{\mathcal{O}}\left(\mathbf{T}^w_{b_k}, \mathbf{T}^w_{b_l}, \mathbf{\hat T}^{b_k}_{b_l}\right)=
    \begin{bmatrix}
    ({\mathbf{R}^w_{b_k}})^T \left( \mathbf{p}^w_{b_l} - \mathbf{p}^w_{b_k} \right) - \mathbf{\hat p}^{b_k}_{b_l} \\
    (\mathbf{\hat R}^{b_k}_{b_l})^T (\mathbf{R}^w_{b_k})^T \mathbf{R}^w_{b_l}
    \end{bmatrix}
    \label{eq:OdomPoseResidual}    
\end{aligned}
\end{equation}
where $b_k$ and $b_l$ are two adjacent keyframes. Within these residuals, the full PGO problem is formulated as:
\begin{equation} 
\begin{aligned}
    \mathop{\mathrm{min}}\limits_{\mathbf{T}^w_b,\mathbf{T}^w_s} 
    & \sum_{(b_k, b_l)\in \mathcal{O}_b} \rho( \lVert \mathbf{r}_{\mathcal{O}}\left(\mathbf{T}^w_{b_k}, \mathbf{T}^w_{b_{l}}, \mathbf{\hat T}^{b_{k}}_{b_{l}}\right) \rVert_{\mathbf{\Lambda}_{\mathcal{O}}}^{2}) + \\ &
    \sum_{(s_k, s_l)\in \mathcal{O}_s} \rho( \lVert \mathbf{r}_{\mathcal{O}}\left(\mathbf{T}^w_{s_k}, \mathbf{T}^w_{s_{l}}, \mathbf{\hat T}^{s_{k}}_{s_{l}}\right) \rVert_{\mathbf{\Lambda}_{\mathcal{O}}}^{2}) + \\ &
    \sum_{(b_i, s_i)\in \mathcal{L}_{\mathcal{O}}} \rho( \lVert \mathbf{r}_{\mathcal{L}_\mathcal{O}}\left(\mathbf{T}^w_{b_i}, \mathbf{T}^w_{s_{i}}, \mathbf{\hat T}^{b_{i}}_{s_{i}}\right) \rVert_{\mathbf{\Lambda}_{\mathcal{L}_\mathcal{O}}}^{2})
    \label{eq:PGO}
\end{aligned}
\end{equation}
where $\mathcal{O}_b$ and $\mathcal{O}_s$ are the set of all odometry residuals of the base map and submap. $\mathbf{\Lambda}_{\mathcal{O}}$ and $\mathbf{\Lambda}_{\mathcal{L}_\mathcal{O}}$ are tunable parameters to describe the uncertainties. Empirically, $\mathbf{\Lambda}_{\mathcal{O}}$ is determined by the drift rate of applied odometry; $\mathbf{\Lambda}_{\mathcal{L}_\mathcal{O}}$ is insensitive to different datasets or scenarios, hence, a general threshold is acceptable. After solving the PGO, odometry drift will be preliminarily eliminated, and all keyframes of multi-session data will be aligned more closely in the world coordinate. More specifically, We will merge the base map $\mathcal{M}_b$ and submap $\mathcal{M}_s$ at the data structure level, including the set of keyframes, the set of odometry residuals, landmarks, and all observations. The process of merging landmarks will be detailed as follows.



PGO improves mapping precision, i.e., the landmark states, by adjusting the poses. Following this, we perform landmark merging to reduce redundancy by merging landmarks that are close to each other. Specifically, we propose three criteria for merging plane landmarks: a. the angle between the normal vectors of two planes is less than a threshold (set to $5^{\circ}$ in the experiments); b. the perpendicular distance of two planes is less than a threshold (set to $0.2$m in the experiments); c. the two planes must have a large overlap, determined by whether the center of one landmark lies within the radius of another landmark. For line landmark merging, the thresholds for line merging are set as $5^{\circ}$ for the angle between the directional vectors and $1.0 \text{m}$ for the perpendicular distance. The landmark merging significantly reduces the number of landmarks and creates a more compact co-visibility structure.



\subsubsection{Bundle Adjustment}
\label{sec:localBA}



Drifted odometry could result in incorrect map vectorization, leading to inaccurate relative pose estimations between sessions that cannot be fully resolved by PGO. To address this, we propose a BA approach to improve map refinement by jointly adjusting landmarks and poses. Inspired by visual BA~\cite{triggs2000bundle}, we design the residual function between the observation in the keyframe and the corresponding landmark in the world. Geometrically, a line can be defined by two points. Thus, every line landmark has two point-to-infinite-line residuals:
\begin{equation} 
\begin{aligned}
    &\mathbf{r}_{\mathcal{L}}\left( \mathbf{T}_{f}^{w},\mathbf{l}_{\mathcal{L}},\mathbf{f}_\mathcal{L} \right) =\\
    &\left[ \begin{array}{c}
    	\mathbf{R}^T(\alpha ,\beta )_{[2\times3]}\left( \mathbf{I}-\mathbf{nn}^T \right) \left( \mathbf{T}_{f}^{w}\mathbf{\hat p}_{a}-\mathbf{q} \right)\\
	\mathbf{R}^T(\alpha ,\beta )_{[2\times3]}\left( \mathbf{I}-\mathbf{nn}^T \right) \left( \mathbf{T}_{f}^{w}\mathbf{\hat p}_{b}-\mathbf{q} \right)\\
\end{array} \right] \in \mathbb{R} ^4    
\end{aligned}
\label{eq:LineResidual}
\end{equation}
where $\mathbf{T}^w_f$ is the keyframe pose; $\mathbf{l}_{\mathcal{L}}$ is the line landmark and its point-direction form is $\left( \mathbf{n}, \mathbf{q} \right)$; $(\cdot)_{[2\times3]}$ represents the first two rows of the matrix, because we use two-dimensional point-to-line distance other than general form such as $(\mathbf{I}-\mathbf{n}\mathbf{n}^T)(\mathbf{p} - \mathbf{q})$. 



Similarly, a plane can be defined by three non-collinear points. The plane landmark and its observation have three point-to-infinite-plane residuals, formulated as follows,
\begin{equation} 
    \mathbf{r}_{\mathcal{S}}\left( \mathbf{T}_{f}^{w},\mathbf{l}_{\mathcal{S}},\mathbf{f}_\mathcal{S} \right) =\left[ \begin{array}{c}
    	\mathbf{n}^T\mathbf{T}_{f}^{w}\mathbf{\hat p}_{a}+d\\
    \mathbf{n}^T\mathbf{T}_{f}^{w}\mathbf{\hat p}_{b}+d\\
    \mathbf{n}^T\mathbf{T}_{f}^{w}\mathbf{\hat p}_{c}+d\\
\end{array} \right] \in \mathbb{R} ^3
    \label{eq:SurfaceResidual}
\end{equation}
where $\mathbf{l}_{\mathcal{S}}$ is the plane landmark and its distance-normal form is $\left( \mathbf{n}, d \right)$.

\begin{figure}
    \centering
    \includegraphics[width=0.95\linewidth]{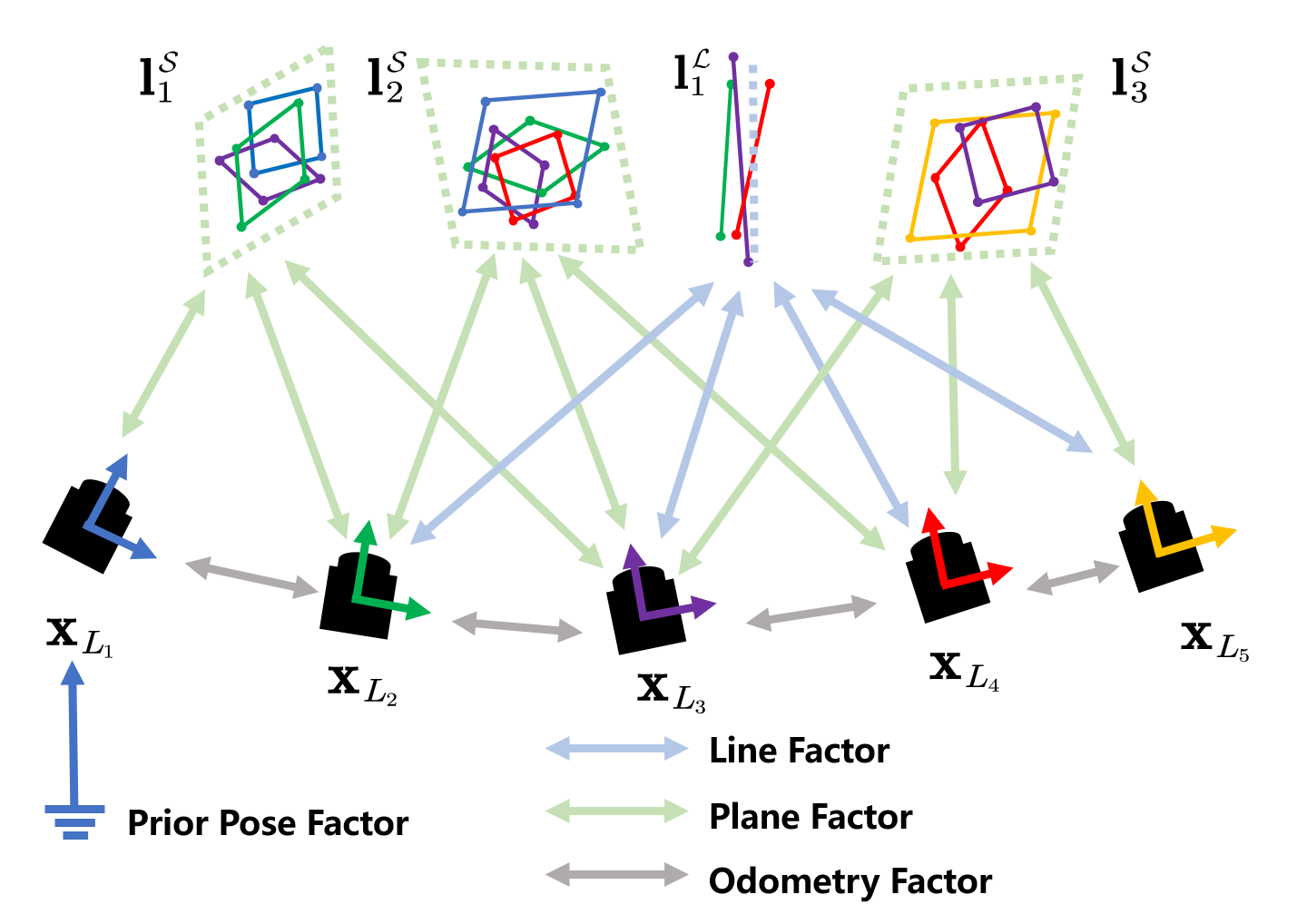}
    \caption{Factor graph of the proposed LiDAR BA. Residuals are constructed from the odometric poses, point-to-line, and point-to-plane measurements. A prior pose factor is utilized to fix the global pose for optimization.}
    \label{fig:BaFactorGraph}
\end{figure}

With the Equations \eqref{eq:LineConvert}, \eqref{eq:SurfConvert}, \eqref{eq:LineResidual} and \eqref{eq:SurfaceResidual}, we can derive the Jacobian matrices with respect to line parameters $\langle\alpha, \beta, x, y\rangle$ and plane parameters $\langle \alpha, \beta, d\rangle$. Then the LiDAR BA can be formulated as follows
\begin{equation}
\begin{aligned}
    \mathop{\mathrm{min}}\limits_{\mathbf{T}^w_f, \mathbf{l}_{\mathcal{L}}, \mathbf{l}_{\mathcal{S}}} 
    \sum_{(f_k, f_l)\in \mathcal{O}} &\rho( \lVert \mathbf{r}_{\mathcal{O}}\left(\mathbf{T}^w_{f_k}, \mathbf{T}^w_{f_l}, \mathbf{\hat T}^{f_k}_{f_l}\right) \rVert_{\mathbf{\Lambda}_{\mathcal{O}}}^{2}) +\\ 
    \sum_{(i, j)\in \mathcal{O}_{\mathcal{L}}} &\rho( \lVert \mathbf{r}_{\mathcal{L}}\left( \mathbf{T}^w_{f_i}, \mathbf{l}_{{\mathcal{L}}_j}, \mathbf{f}_\mathcal{L} \right) \rVert_{\mathbf{\Lambda}_{\mathcal{L}_{i,j}}}^{2}) + \\ 
    \sum_{(i, j)\in \mathcal{O}_{\mathcal{S}}} &\rho( \lVert \mathbf{r}_{\mathcal{S}}\left( \mathbf{T}^w_{f_i},\mathbf{l}_{{\mathcal{S}}_j},\mathbf{f}_\mathcal{S} \right) \rVert_{\mathbf{\Lambda}_{\mathcal{S}_{i,j}}}^{2})
    \label{eq:LBA}
\end{aligned}
\end{equation}
where $\mathcal{O}=\mathcal{O}_b \cup \mathcal{O}_s$ represents the odometry residual set; $\mathcal{O}_\mathcal{L}$ and $\mathcal{O}_\mathcal{S}$ are point-to-line observation set and point-to-plane observation set, respectively. Please refer to the Appendix~\ref{sec:appendix} for the derivations of all Jacobians and more detailed illustrations. We also present a graphical illustration in Figure~\ref{fig:BaFactorGraph} with factor graph representation. We will fix a certain frame to ensure the stability of the optimization. Considering the outliers in block registration caused by odometry drift, we do not integrate cross-session relative poses into the BA optimization.


The overall map refinement is performed in two steps: a pose-only approach (PGO) followed by a pose-landmark joint approach (BA), which offers two key advantages. First, the refinement process is conducted in a coarse-to-fine manner, which enhances the convergence of optimization. Performing BA directly without PGO may lead to convergence failure, as observed during tests using the SLIM system. Second, the optimized poses obtained from PGO serve as an initial guess for BA, thus improving the efficiency of the map refinement. 

To this end, multi-session maps are merged and refined sequentially, as detailed in Sections~\ref{sec:merge} and \ref{sec:refinement}. It is worth noting that these two modules utilize only parameterized lines and planes provided by the map vectorization in Section~\ref{sec:representations}. Using such lightweight representations ensures memory efficiency while maintaining mapping accuracy. The experimental sections will validate the superiority of the proposed modules individually in terms of accuracy, efficiency, and generalizability across different datasets.

%% file: sections/nfr.tex
\section{Map-centric Marginalization}
\label{sec:nfr}

The dimensions of optimization will increase over time, which is a typical issue for long-term mapping. This growth leads to significant time costs for optimization. This section presents a map-centric marginalization that aims to maintain a factor graph with sparsified poses and unaltered landmarks as sessions increase. The proposed map-centric marginalization maintains global consistency while bounding the computational requirements for long-term mapping.





\subsection{Motivation and Problem Formulation}

In classical SLAM systems, \textit{marginalization} is a crucial technique designed to prune the graph and preserve prior information without breaking the observability of the system~\cite{carlevaris2013generic}. The theoretical basis for this technique is the Schur complement~\cite{fan2024schurvins}. However, the Schur complement generally introduces a ``fill-in'' of non-zero blocks in the Hessian matrix, significantly slowing down the nonlinear optimization. Nonlinear factor recovery~\cite{mazuran2014nonlinear,mazuran2016nonlinear} approximates a new distribution of the original problem with fewer recovered factors. NFR is a classical method and plays a key role in the marginalization process of this study. The mathematical description of the NFR problem will be detailed as follows.



\input{figure-caption/nfr_pipeline}

Assuming that all the state variables, i.e., the robot poses and parameterized map landmarks, follow a Gaussian distribution: $p_o\left( \mathbf{s} \right) \sim \mathcal{N}\left( \mu _o,\mathbf{H}_{o}^{-1} \right) $, where $\mathbf{X}$ encodes the state variables; $\mu _o$ and $\mathbf{H}_{o}$ are the current estimation and Hessian matrix (nearly equal to the inverse of covariance), respectively. The goal of NFR is to recover a new distribution $p_r\left( \mathbf{s} \right) \sim \mathcal{N}\left( \mu _r,\mathbf{H}_{r}^{-1} \right)$ to approximate the original distribution $p_o\left( \mathbf{X} \right)$ with a sparse topology. Essentially, this problem is a Kullback-Leibler divergence (KLD) minimization between the original distribution and the recovered distribution, described as follows:
\begin{equation} 
    \begin{aligned}
    &D_{KL}\left( p_o\left( \mathbf{X} \right) ||p_r\left( \mathbf{X} \right) \right) \\
    &=\frac{1}{2}\left( \left< \mathbf{H}_r, \mathbf{H}_{o}^{-1} \right> -\log\det \left( \mathbf{H}_r \right) +\lVert \mathbf{H}_{r}^{\frac{1}{2}}\left( \mu _r-\mu _o \right) \rVert ^2-d \right)         
    \end{aligned}
    \label{eq:KLDMin}
\end{equation}
where $d$ is the dimension of states. 

Furthermore, we define the recovered residual as $\mathbf{r}_k(\mathbf{X}, \mathbf{z}_{r_k})=(f(\mathbf{X}), \mathbf{z}_{r_k}, \mathbf{\Sigma}^{-1}_{r_k})$, where $\mathbf{z}_{r_k}$ is the $k$-th \textit{recovered observation}; $f(\mathbf{X})$ represents the residual function that encompasses $\mathbf{r}_\mathcal{O}$, $\mathbf{r}_{\mathcal{L}}$ and $\mathbf{r}_\mathcal{S}$; $\mathbf{\Sigma}^{-1}_{r_k}$ represents the $k$-th \textit{recovered information matrix}. Since modification of the current state variable estimation is irrational, the mean of the recovered distribution $\mu_r$ must equal to the mean of the original distribution $\mu_o$.  Accordingly, two conditions must be satisfied: $\mathbf{r}_k(\mathbf{\mu}_o, \mathbf{z}_{r_k})=0$ and $\mathbf{\mu}_o=\mathbf{\mu}_r$. Therefore, the original problem in Equation~\eqref{eq:KLDMin} can be reformulated by excluding the third term and the constant fourth term.
\begin{equation} 
\begin{aligned}
    &\mathop{\mathrm{min}}\limits_{\mathbf{\Sigma}_r}\frac{1}{2}\left( \left< \mathbf{J}_{r}^{T}\mathbf{\Sigma }_{r}^{-1}\mathbf{J}_r, \mathbf{H}_{o}^{-1} \right> -\log\det \left( \mathbf{J}_{r}^{T}\mathbf{\Sigma }_{r}^{-1}\mathbf{J}_r \right) \right) 
    \\
    &\text{s}.\text{t}. \mathbf{\Sigma }_{r}^{-1}\succeq 0 \\
    &\mathbf{J}_r=\left[ \begin{array}{c}
	\mathbf{J}_{r_1}\\
	\mathbf{J}_{r_2}\\
	\vdots\\
	\mathbf{J}_{r_K}\\
    \end{array} \right]=\left[ \begin{array}{c}
	\partial \mathbf{r}_1(\mathbf{X}, \mathbf{z}_{r_1})/\partial \mathbf{X}\\
	\partial \mathbf{r}_2(\mathbf{X}, \mathbf{z}_{r_2})/\partial \mathbf{X}\\
	\vdots\\
	\partial \mathbf{r}_K(\mathbf{X}, \mathbf{z}_{r_K})/\partial \mathbf{X}\\
    \end{array} \right] \\
    &\mathbf{\Sigma }_r = \text{diag}(\mathbf{\Sigma }_{r_1},\mathbf{\Sigma }_{r_2},\cdots,\mathbf{\Sigma }_{r_K})
\end{aligned}
\label{eq:KLDMinReForm}
\end{equation}
where $\mathbf{J}_r$ is all the concatenated jacobian blocks and $\mathbf{\Sigma}_r$ denotes the covariance matrix, which is the inverse of the information matrix to be estimated. It should be noted that if linearization point \(\mathcal{X}\) and the observations \(\mathbf{z}_{r_k}\) are known, \(\mathbf{J}_r\) will be a constant matrix. Furthermore, if \(\mathbf{J}_r\) is invertible, a closed-form solution exists~\cite{hsiung2018information,mazuran2016nonlinear}:
\begin{equation}
    \mathbf{\Lambda }_{r_k}=\mathbf{\Sigma }_{r_k}^{-1}=\,\,\left\{ \left( \mathbf{J}_r\mathbf{H}_{o}^{-1}\mathbf{J}_{r}^{T} \right) ^{\left( k \right)} \right\} ^{-1}
    \label{eq:NFRSolution}
\end{equation}


Figure~\ref{fig:nfr_pipeline} presents a graphical illustration of NFR and its corresponding graph topology. Specifically, our map-centric marginalization encompasses two primary steps, as detailed in the following section. First, a new topology is reconstructed from the original by keyframe removal; second, the information matrices of these residuals, which also serve as the solutions to the KLD problem, will be calculated as described in Equation \eqref{eq:NFRSolution}.


\subsection{Two-step Marginalization}
\label{sec:marg}

\subsubsection{Topology Reconstruction}
\label{sec:TopologyReconstruction}
Firstly, we need to select keyframes for marginalization to obtain a new factor graph topology. A density-based downsampling method is applied to remove some keyframe poses in factor graphs. Specifically, a distance threshold is used to control the density of keyframe poses. To ensure the invertibility of the Jacobian matrix in Equation~\eqref{eq:KLDMinReForm}, we propose the following criteria for connectivity specification:
\begin{enumerate}
    \item Each landmark is connected only to the nearest keyframe pose.
    \item All retained keyframes are connected by a minimum spanning tree, and $N-1$ keyframe-to-keyframe residuals are preserved ($N$ is the number of retained keyframes).
    \item Only one unique prior keyframe residual is needed.
\end{enumerate} 
\par The reconstructed Jacobian matrix is always square and invertible, meeting the requirements for Equation~\eqref{eq:NFRSolution}. This is due to two reasons: the dimension of all keyframe-to-landmark residuals always equals the dimension of landmark states; the sum of the dimensions of keyframe-to-keyframe residuals and the dimension of the prior keyframe pose residual exactly equals the dimension of retained keyframe states. It is worth noting that the topology reconstruction is performed after map refinement in Section~\ref{sec:refinement}. Therefore, even though the minimum spanning tree no longer includes any loop closures, the topology reconstruction does not affect the utilization of loop closures for map refinement.





Keyframe removal leads to information loss. Therefore, marginalization is needed to maintain the necessary information for mapping. The classical Schur-complement-based marginalization is formulated as follows. Assume that we have $M$ line landmarks, $N$ plane landmarks, $P$ retained keyframes and $Q$ marginalized keyframes. The dimensions are: retained state variables $r= M\times4 + N\times3 + P\times6$; landmarks $l=M\times4 + N\times3$; retained frame $n=P\times6$; marginalized state variables $m=Q\times6$. Schur-complement-based marginalization is conducted when redundant keyframes are selected, providing the marginalized prior Hessian matrix. Given the Hessian matrix before marginalization $\mathbf{H}$, the covariance matrix $\mathbf{\Sigma }_{rr}^{m}$ after marginalization is derived as:
\begin{equation}
    \begin{aligned}
    \mathbf{H}&=\left[ \begin{matrix}
    	\mathbf{H}_{rr}&		\mathbf{H}_{rm}\\
    	\mathbf{H}_{rm}^{T}&		\mathbf{H}_{mm}\\
    \end{matrix} \right] \in \mathbb{R} ^{\left( r+m \right) \times \left( r+m \right)}  \\ 
    \mathbf{\Sigma }_{rr}^{m}&=(\mathbf{H}_{rr}^{m})^{-1}=(\mathbf{H}_{rr}-\mathbf{H}_{rm}\mathbf{H}_{mm}^{-1}\mathbf{H}_{rm}^{T})^{-1} \\
    &=\left[ \begin{matrix}
    	\mathbf{\Sigma }_{ll}^{m}&		\mathbf{\Sigma }_{ln}^{m}\\
    	{\mathbf{\Sigma }_{ln}^{m}}^T&		\mathbf{\Sigma }_{nn}^{m}\\
    \end{matrix} \right] \in \mathbb{R} ^{r\times r}
    \end{aligned}
    \label{eq:NFRDirectSolution}
\end{equation}
where $\mathbf{H}^m_{rr}$ is the Hessian matrix after the Schur complement, which refers to the $\mathbf{H}_o$ in Equation~\eqref{eq:KLDMinReForm}. It is worth noting that $\mathbf{H}^m_{rr}$ no longer maintains its sparse property due to the Shur complement, and its inverse $\mathbf{\Sigma}^m_{rr}$ is completely dense. The marginalization aims to compute $\mathbf{\Sigma}^m_{rr}$, i.e., the matrix $\mathbf{H}_o^{-1}$, and determine all $\mathbf{\Lambda }_{r_i}$ in Equation~\eqref{eq:NFRSolution}.

Marginalization with the Schur complement generally prunes states that include both landmarks and keyframes, as seen in previous visual mapping by Usenko et al.~\cite{usenko2019visual}. A relatively small dimension of state variables is retained in the marginalization. However, in this study, providing high-quality and complete landmarks is the goal for the SLIM system, and a map-centric marginalization approach is desired to maintain all map landmarks. Unlike previous works, our variable dimensions will range from hundreds of thousands to millions. The excessively large problem dimension leads to dual challenges of slow computation speed and insufficient memory when directly solving Equation \eqref{eq:NFRDirectSolution}. This necessitates leveraging the inherent sparsity of the BA problem to achieve an efficient solution.

To address this challenge, we propose an \textit{equivalent solution via sparse matrix operations}, which significantly improves the efficiency of marginalization and reduces memory consumption, making the SLIM system easily deployable for long-term operations. The relevant sparse matrix operations for efficient NFR will be detailed in the following section.




\subsubsection{Efficient NFR}
\label{sec:EfficientNFR}



The purpose of efficient NFR is to avoid directly calculating $\mathbf{\Sigma}^m_{rr}$ and instead use the sparse matrix blocks of the Hessian matrix before marginalization $\mathbf{H}$ to accelerate the computation of Equation \eqref{eq:NFRSolution}. As previously mentioned, $\mathbf{\Sigma}^m_{rr}$ exhibits both high dimensionality and a dense structure. However, the sparsity inherent in the designed topology obviates the need to compute all blocks in $\mathbf{\Sigma}^m_{rr}$. For instance, the residual associated with a keyframe-to-landmark measurement with index $k$ and the corresponding Jacobian matrix is
\begin{equation}
\begin{aligned}
    \mathbf{J}_{r_k}&=\left[ 0,\cdots , \mathbf{J}_{\mathbf{l}_i}^{k},0,\cdots ,0,\mathbf{J}_{\mathbf{T}_j}^{k},\cdots ,0 \right] 
\end{aligned}
\end{equation}

Given the covariance matrix $\mathbf{\Sigma}^m_{rr}$, the recovered information matrix is derived as 
\begin{equation}
\begin{aligned}
    &\mathbf{\Lambda }_{r_k}=\mathbf{J}_{r_k}\mathbf{\Sigma }_{rr}^{m}\mathbf{J}_{{r_k}}^{T}
    \\
    &=\mathbf{J}_{\mathbf{l}_i}^{k}\mathbf{\Sigma }_{\mathbf{l}_i\mathbf{l}_i}^{m}\left( \mathbf{J}_{\mathbf{l}_i}^{k} \right) ^T+\mathbf{J}_{\mathbf{l}_i}^{k}\mathbf{\Sigma }_{\mathbf{l}_i\mathbf{T}_j}^{m}\left( \mathbf{J}_{\mathbf{T}_j}^{k} \right) ^T+\\
    &\mathbf{J}_{\mathbf{T}_j}^{k}\left( \mathbf{\Sigma }_{\mathbf{l}_i\mathbf{T}_j}^{m} \right) ^T\left( \mathbf{J}_{\mathbf{l}_i}^{k} \right) ^T+\mathbf{J}_{\mathbf{T}_j}^{k}\mathbf{\Sigma }_{\mathbf{T}_j\mathbf{T}_j}^{m}\left( \mathbf{J}_{\mathbf{T}_j}^{k} \right) ^T \\
\end{aligned}
\label{eq:NFRDetailedSolution}
\end{equation}

We can observe that the involved blocks are $\mathbf{\Sigma }_{\mathbf{l}_i\mathbf{l}_i}^{m}$ in $\mathbf{\Sigma}^m_{ll}$, $\mathbf{\Sigma }_{\mathbf{l}_i\mathbf{T}_j}^{m}$ in $\mathbf{\Sigma}^m_{ln}$ and $\mathbf{\Sigma }_{\mathbf{T}_j\mathbf{T}_j}^{m}$ in $\mathbf{\Sigma}^m_{nn}$. Notably, all the $\mathbf{\Sigma }_{\mathbf{l}_i\mathbf{l}_i}^{m}$ are diagonal blocks of $\mathbf{\Sigma}^m_{ll}$, so the off-diagonal blocks of $\mathbf{\Sigma}^m_{ll}$ do not need to be computed. Therefore, the essential blocks are the diagonal blocks of $\mathbf{\Sigma}^m_{ll}$, $\mathbf{\Sigma}^m_{ln}$ and $\mathbf{\Sigma}^m_{nn}$. Generally, we have $l\gg m>n$ considering the landmarks and keyframes in mapping. Therefore, 
the time and memory consumption of calculation the entire $\mathbf{\Sigma}^m_{ll}$ matrix is dominant, which is precisely what our sparse solution aims to reduce.


We leverage two key identities derived from the Sherman-Morrison-Woodbury formula~\cite{woodbury1950inverting} to achieve both sparsity and computational efficiency. Specifically, for the inverse of matrix addition and partitioned matrix, we can obtain the following derivations in Equation~\eqref{eq:SMWInverseFormula}:
\begin{subequations}
    \begin{align}
        (\mathbf{A}-\mathbf{U}\mathbf{C}\mathbf{V}^T)^{-1}&=\mathbf{A}^{-1}+ \notag \\
        \mathbf{A}^{-1}\mathbf{U}(\mathbf{C}^{-1}&-\mathbf{V}^T\mathbf{A}^{-1}\mathbf{U})^{-1}\mathbf{U}^T\mathbf{A}^{-1} \\
        \begin{bmatrix}
            \mathbf{A}& \mathbf{U} \\ \mathbf{V}^T& \mathbf{C}
        \end{bmatrix}^{-1}&=\begin{bmatrix}
            \mathbf{X}& \mathbf{F} \\ \mathbf{G}& \mathbf{Y}
        \end{bmatrix}    \\
        \mathbf{X} = (\mathbf{A}-\mathbf{U}\mathbf{C}^{-1}&\mathbf{V}^T)^{-1} \\
                = \mathbf{A}^{-1}+\mathbf{A}^{-1}&\mathbf{U}(\mathbf{C}-\mathbf{V}^T\mathbf{A}^{-1}\mathbf{U})^{-1}\mathbf{U}^T\mathbf{A}^{-1} \\
        \mathbf{Y} = (\mathbf{C}-&\mathbf{V}^T\mathbf{A}^{-1}\mathbf{U})^{-1} \\
        \mathbf{F} = -\mathbf{A}^{-1}&\mathbf{U}(\mathbf{C}-\mathbf{V}^T\mathbf{A}^{-1}\mathbf{U})^{-1} \\
        \mathbf{G} = -(\mathbf{C}-&\mathbf{V}^T\mathbf{A}^{-1}\mathbf{U})^{-1}\mathbf{U}^T\mathbf{A}^{-1}
    \end{align}
    \label{eq:SMWInverseFormula}
\end{subequations}
then, combing with Equation~\eqref{eq:NFRDirectSolution} and \eqref{eq:SMWInverseFormula}, we have:

\begin{equation} 
\mathbf{\Sigma}^m_{rr}=\mathbf{H}_{rr}^{-1}+\mathbf{H}_{rr}^{-1}\mathbf{H}_{rm}\left( \mathbf{H}_{mm}-\mathbf{H}_{rm}^{T}\mathbf{H}_{rr}^{-1}\mathbf{H}_{rm} \right) ^{-1}\mathbf{H}_{rm}^{T}\mathbf{H}_{rr}^{-1}
\label{eq:NFRSimpleForm}
\end{equation}

\input{figure-caption/hessian_blocks}

In our implementation, we strategically arrange the Hessian blocks, positioning all landmark-related blocks in the top left corner, marginalized blocks (frames) in the bottom right corner, and retained blocks in the center. This arrangement partitions the full Hessian matrix into nine distinct blocks, enabling us to reformulate the original problem presented in Equation~\eqref{eq:NFRDirectSolution} as Equations \eqref{eq:BlockHessian}, \eqref{eq:BlockHessianRR} and \eqref{eq:BlockHessianRM}. As depicted in Figure \ref{fig:nfr_pipeline}, $\mathbf{H}_{ln}$, $\mathbf{H}_{nn}$, $\mathbf{V}_l$, $\mathbf{V}_n$ are sparse matrices and $\mathbf{H}_{ll}$ is a block-diagonal matrix. Figure~\ref{fig:hessian_blocks} presents a graphical understanding of the nine blocks.
\vspace{-25pt} 
\begin{center}
\begin{equation}
    \mathbf{H}= \begin{bmatrix}
    \mathbf{H}_{ll}&		\mathbf{H}_{ln}&		\mathbf{V}_l\\
    (\mathbf{H}_{ln})^T&		\mathbf{H}_{nn}&		\mathbf{V}_n\\
    (\mathbf{V}_{l})^{T}&		(\mathbf{V}_{n})^{T}&		\mathbf{H}_{mm}\\
    \end{bmatrix} \in \mathbb{R}^{\left( r+m \right) \times \left( r+m \right)}
    \label{eq:BlockHessian}
\end{equation}
\end{center}
\vspace{-18pt} 
\begin{center}
\begin{equation}
\mathbf{H}_{rr}= \begin{bmatrix}
    \mathbf{H}_{ll}&		\mathbf{H}_{ln}\\
    (\mathbf{H}_{ln})^T&		\mathbf{H}_{nn}\\
    \end{bmatrix} \in \mathbb{R}^{r\times r}
    \label{eq:BlockHessianRR}
\end{equation}
\end{center}
\vspace{-18pt} 
\begin{center}
\begin{equation}
\mathbf{H}_{rm}= \begin{bmatrix}
    \mathbf{V}_l\\
    \mathbf{V}_n\\
    \end{bmatrix} \in \mathbb{R}^{r\times m}
    \label{eq:BlockHessianRM}
\end{equation}
\end{center}

From Equation~\eqref{eq:NFRSimpleForm}, we identify two crucial matrices: $\mathbf{H}_{rr}^{-1}$ and $\mathbf{R}=\left( \mathbf{H}_{mm}-(\mathbf{H}_{rm})^{T}(\mathbf{H}_{rr})^{-1}\mathbf{H}_{rm} \right)^{-1}$. The block-diagonal property of $\mathbf{H}_{ll}$ allows for the efficient parallel computation of its inverse, which is directly relevant to calculating $\mathbf{H}{rr}^{-1}$. Similarly, the computation of $\mathbf{R}$ can be streamlined by exploiting the $\mathbf{H}_{ll}$ structure. In the subsequent discussion, we will outline a procedure to obtain $\mathbf{R}$ using a series of sparse matrix operations, ensuring low memory consumption and high computational speed.


Let us define $\mathbf{D}=(\mathbf{H}_{ll})^{-1}$ (block-diagonal and positive definite), $\mathbf{P} = (\mathbf{H}_{ll})^{-1}\mathbf{H}_{ln}$ (sparse) and $\mathbf{Q}=(\mathbf{H}_{nn}-(\mathbf{H}_{ln})^{T}(\mathbf{H}_{ll})^{-1}\mathbf{H}_{ln})^{-1}$ (dense and positive definite). We can derive more compact forms to express $\mathbf{H}_{rr}^{-1}$ (block-diagonal and positive definite) and $\mathbf{\Sigma}^m_{rr}$ (dense and positive definite), as follows: 
\begin{equation}
\mathbf{H}_{rr}^{-1}=\begin{bmatrix}
\mathbf{D}+\mathbf{PQP}^T&		-\mathbf{PQ}\\
-\mathbf{Q}^T\mathbf{P}^T&		\mathbf{Q}\\
\end{bmatrix} 
\end{equation}
\begin{equation}
\mathbf{\Sigma }^m_{rr}=\mathbf{H}_{rr}^{-1}+\mathbf{H}_{rr}^{-1}
\begin{bmatrix}
	\mathbf{V}_l\\
	\mathbf{V}_n\\
\end{bmatrix} 
\mathbf{R}\left[ \mathbf{V}_{l}^{T}\,\,\mathbf{V}_{n}^{T} \right] \mathbf{H}_{rr}^{-1}
\end{equation}

Given the sparsity of matrices $\mathbf{V}_l$ and $\mathbf{V}_n$, we can efficiently compute the following \textit{intermediary matrices}: $\mathbf{P}\mathbf{Q}$, $\mathbf{D}\mathbf{V}_l$, $\mathbf{P}^T\mathbf{V}_l$, $\mathbf{P}\mathbf{Q}\mathbf{V}_n$, and $\mathbf{Q}\mathbf{V}_n$. Therefore, computation of $\mathbf{R}$ will involve several sparse matrix operations, as follows:
\begin{equation}
\begin{aligned}
    &\mathbf{R}^{-1}=\mathbf{H}_{mm}-(\mathbf{H}_{rm})^{T}(\mathbf{H}_{rr})^{-1}\mathbf{H}_{rm}\in \mathbb{R} ^{m\times m} \\
    &=\mathbf{H}_{mm} - \left[ \mathbf{V}_{l}^{T}\,\,\mathbf{V}_{n}^{T} \right]\begin{bmatrix}
    \mathbf{D}+\mathbf{PQP}^T&		-\mathbf{PQ}\\
    -\mathbf{Q}^T\mathbf{P}^T&		\mathbf{Q}\\
    \end{bmatrix} \left[ \begin{array}{c}
    \mathbf{V}_l\\
    \mathbf{V}_n\\
    \end{array} \right] \\
    &=\mathbf{H}_{mm}- \mathbf{V}_{l}^{T}\mathbf{DV}_l-\mathbf{V}_{l}^{T}\mathbf{PQP}^T\mathbf{V}_l \\
    &+\left( \mathbf{V}_{l}^{T}\mathbf{PQV}_n \right) ^T+\left( \mathbf{V}_{l}^{T}\mathbf{PQV}_n \right) -\mathbf{V}_{n}^{T}\mathbf{QV}_n 
\end{aligned}
\end{equation}
which enables high-efficiency computing by leveraging advanced packages such as Eigen~\cite{eigenweb} and cuBLAS~\cite{cuda}.


Based on the Schur complement, $\mathbf{R}$ represents the marginalized covariance matrix of the marginalized frames. The matrix $\mathbf{R}$ is inherently positive definite, ensuring its inverse exists. We leverage QR decomposition to derive $\mathbf{R}$ from its inverse, $\mathbf{R}^{-1}$. Furthermore, we exploit the inherent symmetry in the second term of $\mathbf{\Sigma}^m_{rr}$ in Equation~\eqref{eq:NFRSimpleForm}, which can reduce the computational burden by calculating only one part. Specifically, We apply a Cholesky decomposition on $\mathbf{R}$ to obtain $\mathbf{R}=\mathbf{R}_c(\mathbf{R}_c)^T$, and we have
\begin{equation} 
\begin{aligned}
    \mathbf{W}&=\begin{bmatrix}
    \mathbf{D}+\mathbf{PQP}^T&		-\mathbf{PQ}\\
    -\mathbf{Q}^T\mathbf{P}^T&		\mathbf{Q}\\
    \end{bmatrix} \begin{bmatrix}
    	\mathbf{V}_l\\
    	\mathbf{V}_n\\
    \end{bmatrix} \mathbf{R}_c\\
    &=\begin{bmatrix}
    	\mathbf{D}\left( \mathbf{V}_l\mathbf{R}_c \right) +\mathbf{PQ}\left[ \left( \mathbf{P}^T\mathbf{V}_l-\mathbf{V}_n \right) \mathbf{R}_c \right]\\
    	-\mathbf{Q}\left[ \left( \mathbf{P}^T\mathbf{V}_l-\mathbf{V}_n \right) \mathbf{R}_c \right]\\
    \end{bmatrix}\\
    \mathbf{W}_l &= \mathbf{D} \mathbf{V}_l\mathbf{R}_c +\mathbf{PQ} \left( \mathbf{P}^T\mathbf{V}_l-\mathbf{V}_n \right) \mathbf{R}_c  \\
    \mathbf{W}_n &= -\mathbf{Q} \left( \mathbf{P}^T\mathbf{V}_l-\mathbf{V}_n \right) \mathbf{R}_c  \\
    \mathbf{\Sigma }^m_{rr}&=\begin{bmatrix}
    \mathbf{D}+\mathbf{PQP}^T&		-\mathbf{PQ}\\
    -\mathbf{Q}^T\mathbf{P}^T&		\mathbf{Q}\\
    \end{bmatrix}+\begin{bmatrix}
        \mathbf{W}_l\mathbf{W}_l^T&    \mathbf{W}_l\mathbf{W}_n^T \\
        \mathbf{W}_n\mathbf{W}_l^T&    \mathbf{W}_n\mathbf{W}_n^T \\
    \end{bmatrix}
\end{aligned}    
\end{equation}
in which \textit{intermediary matrices} in these equations can be re-used directly as intermediate results. 

As aforementioned, we only need to compute $\mathbf{\Sigma}_{ln}^m= -\mathbf{P}\mathbf{Q} + \mathbf{W}_l\mathbf{W}_n^T$ and $\mathbf{\Sigma}_{nn}^m=\mathbf{Q} + \mathbf{W}_n\mathbf{W}_n^T$ and the diagonal blocks of $\mathbf{\Sigma}_{ll}^m=\mathbf{D} + \mathbf{P}\mathbf{Q}\mathbf{P}^T+\mathbf{W}_l\mathbf{W}_l^T$. Notably, the computation of the $\mathbf{D} + \mathbf{P}\mathbf{Q}\mathbf{P}^T+\mathbf{W}_l\mathbf{W}_l^T$ can be accelerated through parallel processing, further enhancing efficiency. The remaining two matrices, characterized by their lower dimensionality, can be computed directly. The situation we discussed earlier only considers the Jacobian matrix for keyframe-to-landmark residual. The Jacobian matrices for keyframe-to-keyframe and prior pose measurements depend solely on $\mathbf{Q} + \mathbf{W}_n\mathbf{W}_n^T$, ensuring the feasibility of our proposed sparse solution. 

Finally, within the each single Jacobian block $\mathbf{J}_{r_k}$ and the essential blocks of $\mathbf{\Sigma}^m_{rr}$, i.e. $\mathbf{H}_o^{-1}$, all the recovered information matrices $\left\{\mathbf{\Lambda}_{r_k} \right\}$ could be calculated with Equation \eqref{eq:NFRDetailedSolution} in parallel. So far, we have introduced the construction of all recovered observations after marginalization, resulting in a sparse factor graph that is very close to the dense factor graph. The experimental section will demonstrate the necessity of map-centric marginalization and the effectiveness of our proposed NFR.



%% file: figure-caption/nfr_pipeline.tex
\begin{figure*}[t]
  \centering
  \includegraphics[width=0.95\linewidth]{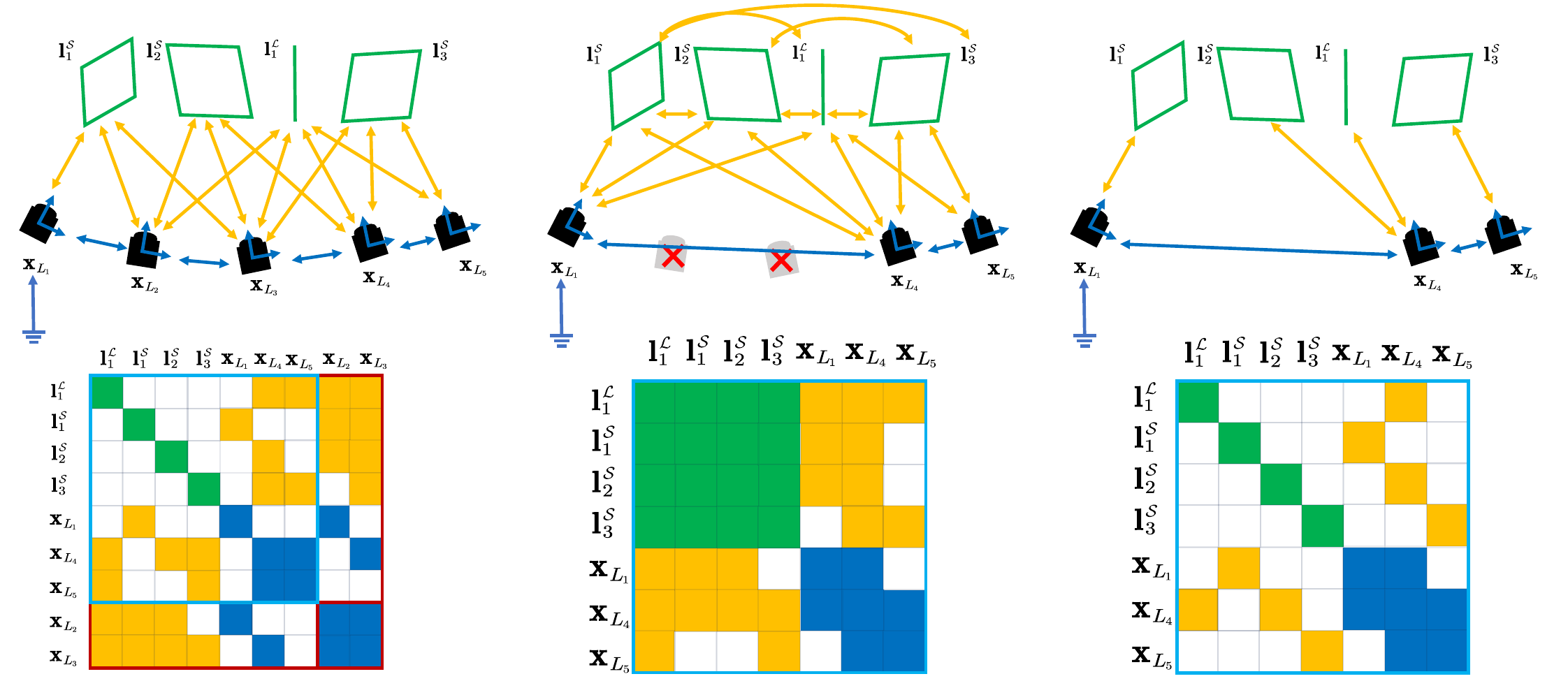}
  \caption{Factor graph topology and its corresponding Hessian matrix. The left part shows the original factor graph and its Hessian matrix, where the green diagonal blocks represent the Hessian matrix for the landmarks. The light-blue box contains all the variables to be retained, while the red box contains the variables to be marginalized. The middle part shows the marginalized factor graph and the dense Hessian matrix, where the landmark part is no longer block-diagonal. Map-centric NFR aims to find the Hessian matrix on the right side. Our solution closely maintains the block diagonal structure; meanwile approximates the probability distribution of the state variables represented by the intermediate problem, using only the sparse matrix block on the left side for calculations.}
  \label{fig:nfr_pipeline}
\end{figure*}


%% file: figure-caption/hessian_blocks.tex
\begin{figure}[t]
  \centering
    \includegraphics[width=0.5\linewidth]{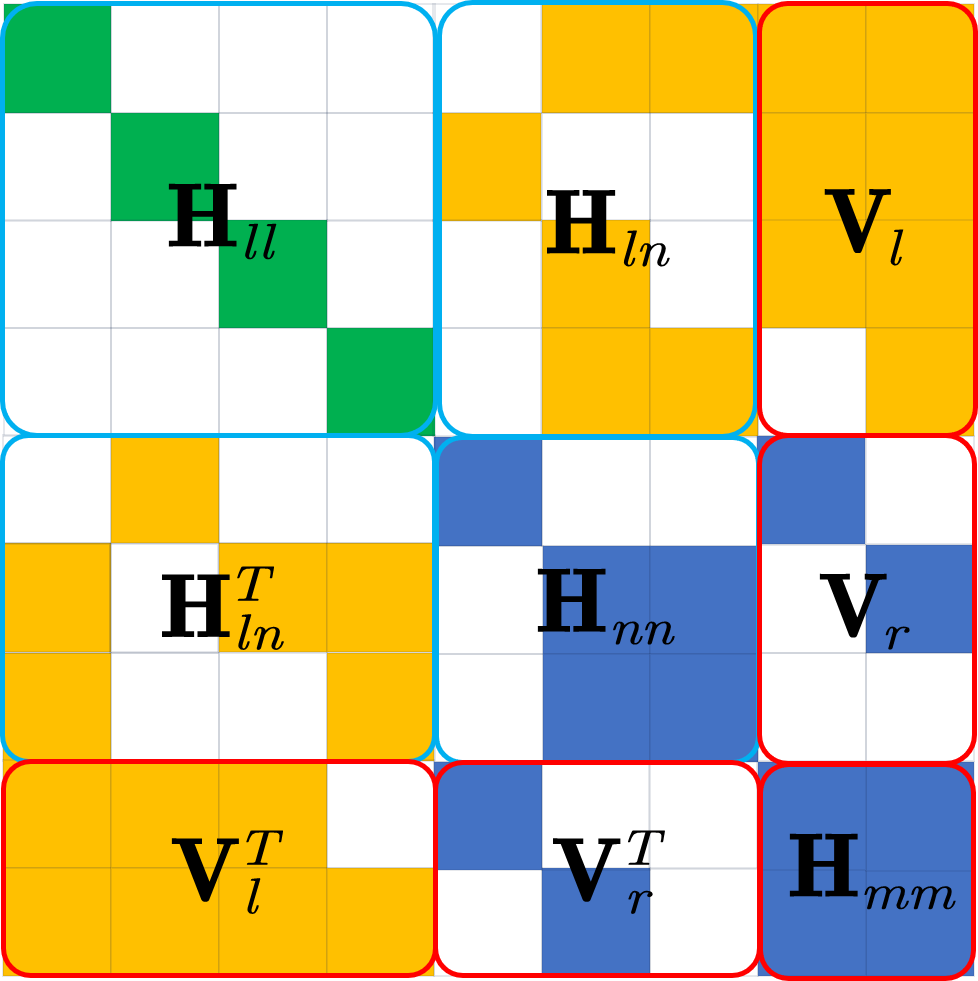}
  \caption{Visualization for the Hessian matrix before marginalization. Noted that $\mathbf{H}_{ll}$ is block-diagonal and all of other blocks are sparse matrices in general. $\mathbf{H}_{ll}$ is much larger in dimensions than $\mathbf{H}_{nn}$ and $\mathbf{H}_{mm}$. Therefore, in the implementation, we only store its diagonal blocks, while all other matrices are saved using a sparse matrices.
  }
  \label{fig:hessian_blocks}
\end{figure}

%% file: sections/experiments.tex
\section{Experiments}
\label{sec:exp}


We conduct real-world experiments to evaluate the performance of the proposed SLIM system. We first introduce the experimental setup in Section~\ref{sec:setup}. The following tests are designed to validate the four capabilities of SLIM and to address the motivations outlined in Section~\ref{sec:introduction}:
\begin{enumerate}
    \item \textit{Accuracy}. Accuracy is the main concern for robotic localization and mapping. The accuracy of SLIM is quantitatively evaluated in Section~\ref{sec:accuracy}, with comparisons to other advanced methods.
    \item \textit{Lightweightness}. In Section~\ref{sec:lightweight}, we compare the requirements for storage space (memory consumption) with other map representations.
    \item \textit{Scalability}. The designed marginalization is validated in Section~\ref{sec:scalability} individually with multi-session maps.
    \item \textit{Localizability}. Section~\ref{sec:localization} demonstrates that the maps from SLIM can be re-used for online robot localization.
\end{enumerate}

\subsection{Experimental Set-up}
\label{sec:setup}

\input{figure-caption/sensors}
\input{figure-caption/nclt_dataset_setting}


Three real-world datasets are employed for comprehensive validation and quantitative evaluation: KITTI~\cite{geiger2012we}, NCLT~\cite{carlevaris2016university}, and HeLiPR~\cite{jung2023helipr}. The KITTI dataset is collected by a vehicle equipped with a Velodyne HDL-64E. We use Sequence 00-10 for experimental validation, although some sequences are not entirely in urban environments. We employ KISS-ICP~\cite{vizzo2023kiss} to obtain LiDAR odometry as prior poses. Each sequence is partitioned into multiple sessions with overlapping regions before proceeding with our map merging pipeline. NCLT is also a well-known dataset suitable for the evaluation of multi-session mapping~\cite{carlevaris2014generic}. Multiple sessions are collected from a mobile platform equipped with a Velodyne HDL-32E. Due to the rotational motion of the platform, we use FAST-LIO2~\cite{xu2022fast} for point cloud undistortion and LiDAR odometry generation. We selected 6 regions as our benchmark, with each region containing between 25 and 60 sessions, meaning the robot travels to the location more than 25 times.

Both KITTI and NCLT have nearly a decade of history. For the recent HeLiPR~\cite{jung2023helipr}, similar to the NCLT dataset, we partition 32 overlapping regions as our benchmark, with each region having around 10 sessions. A fast G-ICP~\cite{koide2021voxelized} is employed as LiDAR odometry on the HeLiPR dataset. The reason for using various LiDAR odometry methods is that LiDAR sensing varies across the three datasets, as shown in Figure~\ref{fig:sensors}. Thus, we select a suitable odometry configuration for each dataset. Overall, these datasets are collected in different countries worldwide and on different platforms, thus verifying the generalization ability of the proposed SLIM system.


As mentioned above, we partition the large-scale urban areas into regions for evaluation. Essentially, the \textit{region} represents a type of submap for long-term management, which differs from the concept of \textit{session}. More specifically, the former is in the spatial domain, while the latter is in the temporal domain. We present the visualization results of the multi-session data and the regions in Figure~\ref{fig:regions} for a better illustration. Figure~\ref{fig:mapping_visualization} also shows the mapping results in different regions. For visualization, a plane landmark is represented by a rhombus formed by selecting four points. These four control points are determined by computing the PCA of the weighted sum of the observed points within each landmark.





 

\subsection{Mapping Accuracy}
\label{sec:accuracy}

\input{table-caption/accuracy}

\input{figure-caption/mapping_visualization}


This section validates that SLIM can provide a globally consistent map from multi-session data. We first compare SLIM with other advanced mapping systems on KITTI, using trajectory accuracy for quantitative evaluation. Then, we demonstrate long-term map merging and consistent mapping with incremental map sessions on the HeLiPR dataset.


\subsubsection{Mapping Accuracy on KITTI Dataset}
\label{sec:accuracyKITTI}

Although KITTI covers multiple scenarios beyond urban environments, such as rural and highway, the proposed SLIM system can still provide line and plane maps. Since KISS-ICP lacks loop closure capability, SLIM autonomously searches for overlapping regions of each submap, estimates relative poses, and then performs subsequent PGO and BA. In practical deployments, robots are frequently equipped with additional global localization functions, such as GPS, to facilitate map merging. The integration of such information provides strong global pose priors for the mapping system, effectively reducing the complexities of global map merging.




Regarding the evaluation metrics, we compute the absolute rotation and translation error between the keyframes in our global map and the corresponding frames in the ground truth trajectory. The benchmark results are summarized in Table \ref{tab:kitti_acc}, with partial results from BALM2~\cite{liu2023efficient}. The overall results indicate that the advanced PIN-SLAM~\cite{pan2024pin} achieves the best performance, while BALM2~\cite{liu2023efficient} and proposed SLIM achieve competitive performance on certain sequences. We observe that SLIM performs poorly on Sequences 01 and 09 due to insufficient roadside features, such as streetlights and trees. 


Among these comparisons, only the proposed SLIM, BALM2~\cite{liu2023efficient}, and PLC-LiSLAM~\cite{zhou2022mathcal} achieve feature-level LiDAR BA. BALM2 primarily focuses on state estimation rather than long-term mapping, as it lacks an explicit representation of landmarks. Additionally, BALM2 uses dense points for data association and residual construction and relies on a second-order solver. In contrast, PLC-LiSLAM uses planes, lines, and cylinders as landmarks with explicit representation. However, its landmark representation is not minimal, which necessitates additional design efforts to ensure correct state updates and maintain the sparsity of the nonlinear optimization problem. Our proposed SLIM employs a minimal landmark representation, ensuring that the corresponding optimization naturally exhibits a sparse structure. Furthermore, we design and validate the map merge and marginalization capabilities within the SLIM system, enhancing its overall functionality.


\subsubsection{Incremental Mapping on HeLiPR Dataset}
\label{sec:accuracyHeLiPR}

\input{figure-caption/trajectory_accuracy}

Compared to the KITTI dataset, the recently published HeLiPR dataset is more suitable for validating long-term mapping. This is due to the inter-LiDAR sequences in HeLiPR, which naturally verify map merging and refinement as sessions increase at the same location. Specifically, the HeLiPR dataset encompasses sequences collected in four different scenarios: DCC, KAIST, Roundabout, and Town, with three sequences in each scenario. To obtain multiple submaps with overlapping regions, we selected 32 regions as our benchmark, with each region containing around 10 sessions. The SLIM system sequentially merges multiple submaps onto the base map, achieving incremental mapping from multi-session data. As depicted in Algorithm~\ref{alg:system}, SLIM conducts PGO and BA when a new session is merged, allowing for evaluation of the merged trajectory accuracy. We also compute the trajectory accuracy of the current submap to be merged. The heights of ground truth trajectories are not stable in some sequences of HeLiPR, so we evaluate trajectory accuracy along the $x$ and $y$ only. 


\input{figure-caption/M2DGR}

\input{figure-caption/cloud_compare}

We present the experimental results of 20 sequences quantitatively in Figure \ref{fig:trajectory_accuracy}. The SLIM system can maintain the consistency of the global map as new maps are added, even if the new maps are aligned with poor prior poses. Notably, the error of BA is generally less than PGO and the original LiDAR odometry, demonstrating the effectiveness of the proposed map refinement in Section~\ref{sec:refinement}. We also present qualitative results by applying our proposed BA in Figure~\ref{fig:ba_cloud_compare}, verifying the necessity of using BA for high-consistency mapping. We observe unstable results after applying PGO due to low-quality relative poses from block registration when odometry poses are not accurate enough. In summary, given multi-session maps, the map refinement modules can effectively reduce pose errors, i.e., improve mapping accuracy, by utilizing our proposed map merging and refinement.



One key capability of our framework is map merging, as described in Section~\ref{sec:merge}, which is harder for quantitative evaluation compared to mapping accuracy. All the input submaps in this section are within their own coordinates when generated. The map merge module can align them into one global frame, resulting in the error reduction shown in Figure~\ref{fig:trajectory_accuracy}. The video in the supplementary materials also presents the map merge process to facilitate understanding.

In addition to the KITTI, NCLT and HeLiPR, we also test the SLIM system qualitatively on the M2DGR dataset~\cite{yin2021m2dgr}.We present the qualitative mapping result in Figure~\ref{figure:m2dgr}. The SLIM system achieves map merging and refinement on the multi-session data provided by M2DGR, demonstrating its effectiveness in structural environments.




\subsection{Memory Efficiency}
\label{sec:lightweight}

\input{table-caption/storage}

\input{figure-caption/nclt_storage}

Urban environments cover large spaces and various scenarios, so memory efficiency is a major concern for mobile robots. In this section, we evaluate the memory efficiency of SLIM compared to several conventional mapping systems in terms of storage size (memory consumption). The SLIM system utilizes lines and planes as basic front-end representations. The stored map can also contain graph structures that encode poses and co-visible information. If the map is solely designed for localization without such structures, we denote it with an (L) label for clarity. The first seven methods (from Eigen-Factor~\cite{ferrer2019eigen} to BALM2~\cite{liu2023efficient}) in Table \ref{tab:kitti_acc} utilize point clouds as explicit representation, resulting in dense point clouds as generated output maps.

The memory consumption of different maps is first evaluated on the KITTI dataset. The comparisons include both explicit representations, such as sampled point clouds (with $r$ as the sampling radius), and recent implicit neural representations~\cite{isaacson2023loner,zhong2023shine,pan2024pin,deng2023nerf}. Partial results are from previous studies~\cite{yu2023multi,pan2024pin}, and all are summarized in Table \ref{tab:storage}. Our lightweight map requires significantly less storage than other mapping systems, demonstrating its superiority in memory efficiency for urban environments. Specifically, SLIM (L) only needs less than 0.5 Megabyte (MB) to cover a travel length of over 3 kilometers (km), with a density of approximately 130 KB/km. If the full SLIM map is needed for updates, the system uses less than 15 MB per sequence to store on disks. 


The memory consumption on NCLT and HeLiPR is presented in Figure~\ref{fig:nclt_storage} and Figure~\ref{fig:nfr_compare}, respectively. In Figure~\ref{fig:nclt_storage}, each region requires less than 1 MB to store a full map structure. Similar results are found in Figure~\ref{fig:nfr_compare}. In summary, the built maps require low storage consumption on these two datasets, consistent with the maps generated from KITTI.

It is worth noting that the ``efficiency'' and ``lightweight'' in this study differ from those in recent LiDAR odometry~\cite{xu2022fast,bai2022faster}. The two odometry systems focus on short-term and computationally efficient LiDAR mapping, while the SLIM is designed for long-term and memory-efficient mapping. High memory efficiency could reduce data transmission load in multi-robot systems for distributed operations, such as crowd-sourced urban mapping.


\subsection{Scalability for Long-term Mapping}
\label{sec:scalability}


Sections~\ref{sec:accuracy} and \ref{sec:lightweight} demonstrate that the SLIM system provides not only accurate but also memory-efficient maps in urban environments. The map merge capability is also verified in Section~\ref{sec:accuracy}. As discussed in Section~\ref{sec:introduction}, one motivation for SLIM is scalability, which is crucial for long-term mapping to maintain global consistency and bound map sizes as map sessions increase. Scalability is particularly important to ensure that map refinement, especially for high-dimensional LiDAR BA, remains controllable.

We verify the scalability of SLIM on NCLT and HeLiPR. For NCLT, Figure~\ref{fig:nclt_storage} shows that both map consumption and time cost remain within certain limits even with 50 sessions in the same region. This satisfying performance is primarily attributed to the marginalization module. To verify this, we disable the marginalization and conduct tests on HeLiPR. In Figure~\ref{fig:nfr_compare}, if marginalization is disabled, we observe that map optimization time exhibits near-exponential growth, while map storage consumption increases linearly with the number of sessions. Overall, the full system maintains controllable costs on both NCLT and HeLiPR, verifying the scalability of SLIM. We do not provide experimental results on NCLT, as disabling marginalization will extend submap merging time to tens of minutes or even hours.


The proposed marginalization consists of two steps: topological reconstruction (Section~\ref{sec:TopologyReconstruction}) followed by map-centric NFR (Section~\ref{sec:EfficientNFR}), and the latter is one of the key contributions of this study. One might argue that only topological reconstruction or other advanced keyframe sampling techniques could maintain a sparse map structure for long-term mapping, making the NFR unnecessary due to its time costs in solving Equation~\eqref{eq:KLDMin}. To verify this, we applied topological reconstruction (keyframe sparsification) in our method while maintaining the prior information described in Section~\ref{sec:preprocessing}, i.e., without solving the NFR problem. Specifically, we define this prior information as $\sqrt{\mathbf{\Lambda}_{r_k}} =\mathbf{I} \cdot  \sqrt{N_{r_k}/m}/ \sigma$, considering the sum of points in the original observations, as the number of points directly reflects their weights. Figure~\ref{fig:nfr_prior_compare} indicates that mapping accuracy drops if using priors compared to the full module (NFR enabled). The results demonstrate that NFR plays a vital role in both consistency and scalability for long-term mapping.

The minimum spanning tree generates the sparse topology of keyframe connections. Figure \ref{figure:reduced_graph} shows the effectiveness of our keyframe sparsification approach, ensuring comprehensive connectivity among all keyframes while reducing unnecessary information storage. An additional issue in topology reconstruction is whether the system can preserve mapping accuracy when a landmark is retained with only a single observation on the keyframe, i.e., each landmark is connected only to the nearest keyframe pose, as described in Section~\ref{sec:marg}. Our prior experiments in Figure \ref{fig:trajectory_accuracy} show that the system consistently maintained high accuracy following each BA. The optimized results of the LiDAR BA remain unaffected by the precision of the multi-session input submaps. These results indicate that the proposed map-centric NFR can provide a reasonable information matrix to ensure the stability of mapping refinement, even when each landmark is connected to one keyframe pose after topology reconstruction.



\input{figure-caption/nfr_compare}

\input{figure-caption/nfr_prior_compare}




\input{figure-caption/reduced_graph}

\subsection{Map Re-use for Online Localization}
\label{sec:localization}

SLIM is designed to provide maps that can be reused for online state estimation, i.e., robot localization, a basic need for modern mobile robots. We employ a straightforward LiDAR-based localization algorithm to evaluate whether our generated map can be utilized for robot relocalization. For simplicity, we primarily assess continuous localization capability, while global relocalization can be achieved using the module presented in Section~\ref{sec:graphregistration}. We apply the same feature extraction operations described in Section~\ref{sec:preprocessing} to the original LiDAR scans to obtain features as measurements. Subsequently, we use the pose from the previous frame $\mathbf{T}^w_f$ as the prior pose for the current frame and perform nearest neighbor data association between the points $\mathbf{p}_l$ and landmarks $(\mathbf{n}_k, \mathbf{p}_k)$ of the current frame, thus obtaining the matching pairs $\mathcal{C}^{\mathcal{L}}$ and $\mathcal{C}^{\mathcal{S}}$. Essentially, localization on the map is to minimize the error between observed points and the pre-built maps. We adopt a point-to-landmarks residual to achieve online localization, described as follows:
\vspace{-5pt}
\begin{equation}
    \begin{aligned}
    \mathop{\mathrm{min}}\limits_{\mathbf{T}^w_f}
        & \sum_{(k, l)\in \mathcal{C}^{\mathcal{L}}} \rho( \lVert (\mathbf{I} - \mathbf{n}_k\mathbf{n}^T_k)(\mathbf{T}^w_f \mathbf{p}_l - \mathbf{p}_k) \rVert_{\Lambda_{\mathcal{L}}}^{2}) \\
        & \sum_{(k, l)\in \mathcal{C}^{\mathcal{S}}} \rho( \lVert \mathbf{n}^T_k(\mathbf{T}^w_f \mathbf{p}_l - \mathbf{p}_k) \rVert_{\Lambda_{\mathcal{S}}}^{2})
    \end{aligned}
    \label{eq:OnlineLocalization}
\end{equation}
\vspace{-5pt}

Table~\ref{tab:localization} reports quantitative results on the HeLiPR dataset. The localization error is generally at the centimeter level, which is sufficiently precise for localization-oriented applications, such as autonomous driving in urban environments. Note that the error includes the errors of pre-mapping, resulting in larger errors in certain regions. For practical applications, additional information, such as odometry and semantics~\cite{yu2023multi}, could improve localization accuracy under a sensor fusion scheme. Moreover, the average latency for solving the scan-to-map optimization using a single thread is approximately 42ms (CPU: Intel i7-14650HX), which is sufficient for real-time localization applications.



\input{table-caption/localization}

%% file: figure-caption/sensors.tex
\begin{figure}[t]
  \centering
    \includegraphics[width=0.95\linewidth]{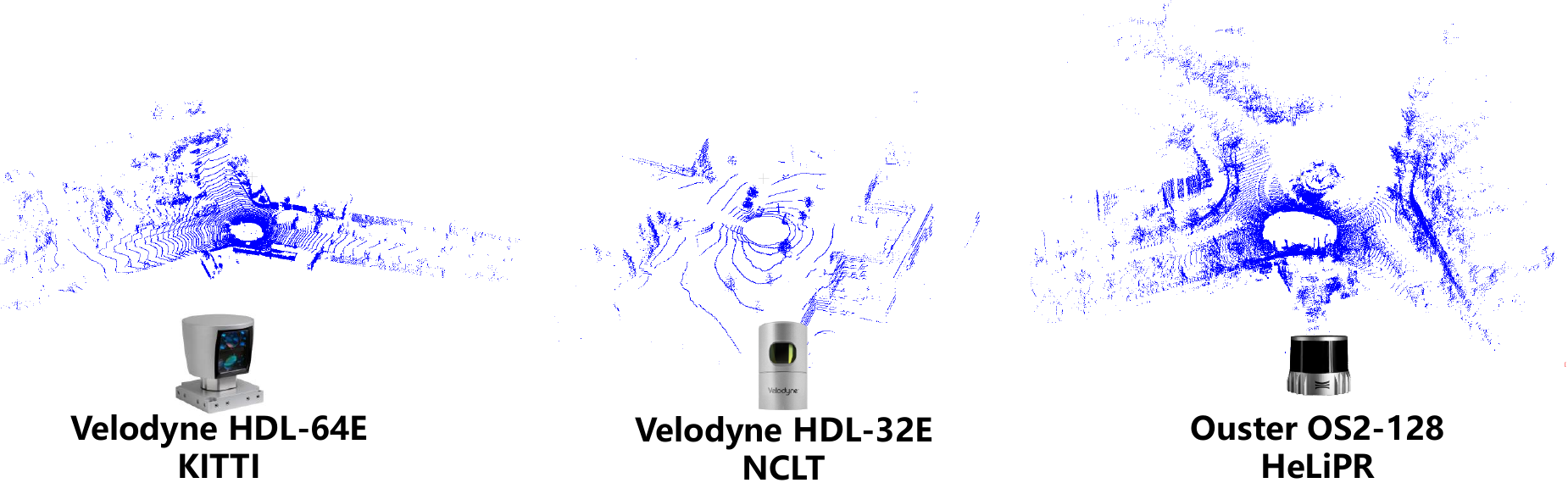}
  \caption{LiDAR sensors and scans in KITTI, NCLT and HeLiPR. The origin of each scan, corresponding to the sensor frame, is positioned at the center of the LiDAR point clouds. The properties of point clouds, such as density, vary with different sensor configurations. The front-end map vectorization module can still parametrize the point clouds to lines and planes, ensuring the functionality of the subsequent map refinement and maintenance modules.}
  \label{fig:sensors}
\end{figure}

%% file: figure-caption/nclt_dataset_setting.tex
\begin{figure}[t]
	\centering
	\subfigure[NCLT]{
        \includegraphics[width=0.95\linewidth]{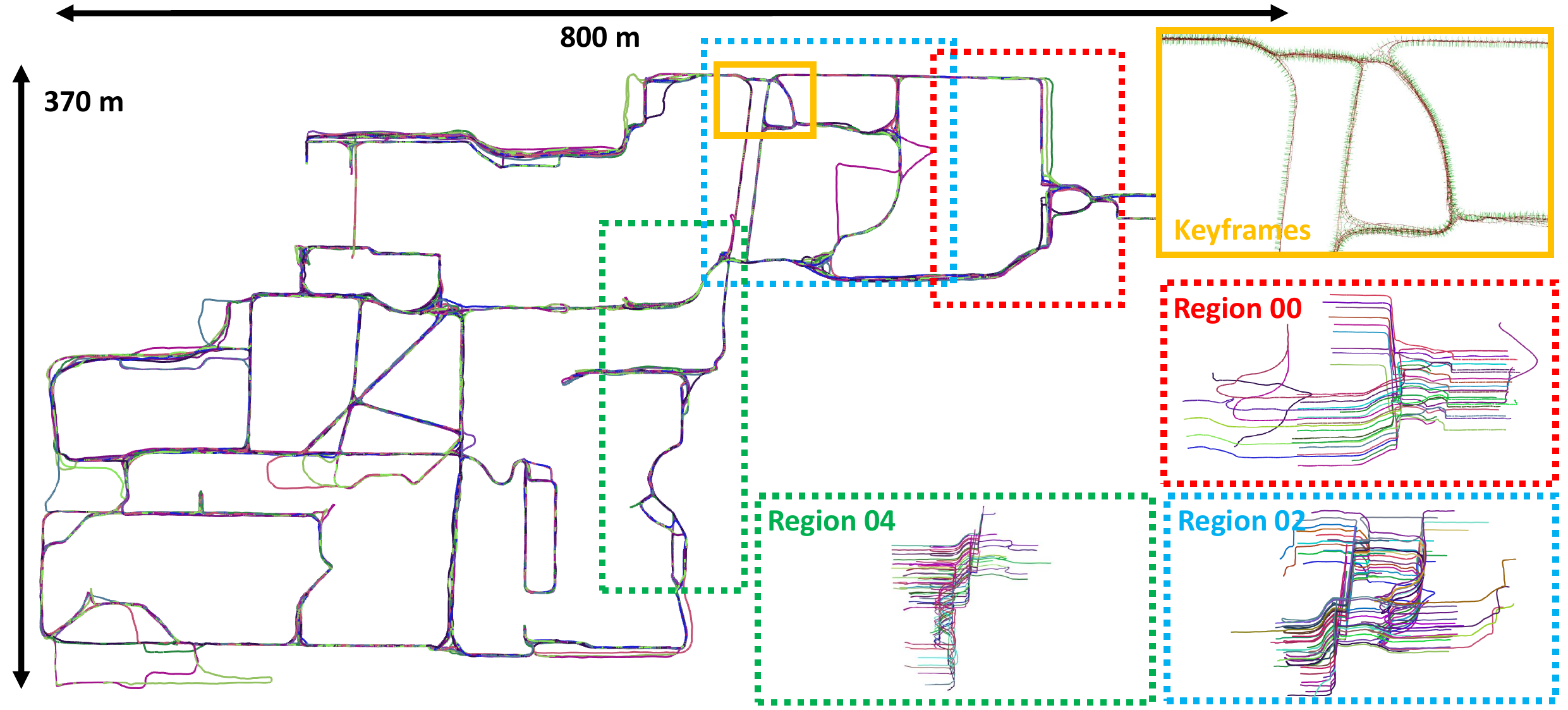}
      }
    \caption{The multi-session datasets in NCLT. We employ the regions, a type of large-scale submaps, to better manage urban mapping. The zoomed-in views in the red, green and blue boxes provide trajectories of multi-session robot travels, visualized with varying heights and colors. The yellow boxes provide zoomed keyframe views of poses on the x-y plane.}
	\label{fig:regions}
\end{figure}

%% file: table-caption/accuracy.tex

\begin{table*}[t]
    \caption{Absolute Trajectory Error (RMSE, Meters) on KITTI}
    \centering
    \renewcommand\arraystretch{1.2}
    \begin{threeparttable}
        {\begin{tabular}{ccccccccccccc}
        \hline
        \hline
        Sequence & 00 & 01 & 02 & 03 & 04 & 05 & 06 & 07 & 08 & 09 & 10 & Mean \\
        \hline
        Eigen-Factor~\cite{ferrer2019eigen} & 1.02 & 1.94 & 5.28 & 0.70 & 0.82 & 0.84 & 0.31 & 0.43 & 2.80 & 1.98 & 0.99 & 1.55\\
        CT-ICP~\cite{dellenbach2022ct} & 1.68 & 2.25 & 4.06 & 0.67 & 0.67 & 0.76 & 0.34 & 0.40 & 2.52 & \textbf{0.91} & 0.83 & 1.40\\
        MULLS~\cite{pan2021mulls} & 1.09 & 1.96 & 5.42 & 0.74 & 0.89 & 0.97 & 0.31 & 0.44 & 2.93 & 2.12 & 1.13 & 1.63\\
        BALM~\cite{liu2021balm} & 0.96& 1.90& 5.21& 0.68& 0.75& 0.72& 0.28& 0.40& 2.72& 1.75& 0.92 & 1.48\\
        Plane Adjustment~\cite{zhou2021pi} &  0.86& 1.84& 5.08& \textbf{0.58}& 0.64& 0.66& 0.23& 0.31& 2.63& 1.50& 0.80 & 1.37\\
        BAREG~\cite{huang2021bundle}     &  0.89& 1.88& 5.12& 0.65& 0.70& 0.69& 0.24& 0.35& 2.68& 1.59& 0.88 & 1.42\\
        BALM2~\cite{liu2023efficient}      &  {0.84}& \textbf{1.83}& 5.06& 0.57& 0.64& 0.62& \textbf{0.21}& \textbf{0.30}& 2.59& 1.48& \textbf{0.78} & 1.34\\
        LONER\footnotemark~\cite{isaacson2023loner} & 14.13 & $\times$ & 69.68 & 5.19 & 1.22 & 6.65 & 1.00 & $\times$ & 17.86 & 6.88 & 7.57 & NA \\
        PIN-LO~\cite{pan2024pin} & 5.6 & 4.3 & 9.3 & 0.7 & \textbf{0.1} & 1.7 & 0.5 & 0.5 & 3.0 & 1.8  & 0.8 & 2.6\\
        PIN-SLAM~\cite{pan2024pin} & \textbf{0.8} & 4.3 & 2.1 & 0.7 & \textbf{0.1} & \textbf{0.3} & 0.4 & \textbf{0.3} & \textbf{2.1} & 1.2  & 0.8 & \textbf{1.2}\\
        SLIM (Ours)      &  0.95& 3.72& \textbf{1.99}& 0.79& 0.28& 0.61& 0.24& 0.33& 2.31 & 3.12& 1.07 & 1.40\\
        \hline
        \hline
        \end{tabular}}
    \end{threeparttable}
    \label{tab:kitti_acc}
\end{table*}

%% file: figure-caption/mapping_visualization.tex
\begin{figure*}[t]
  \centering
  \subfigure[NCLT Region 04]{
    \includegraphics[width=0.48\linewidth]{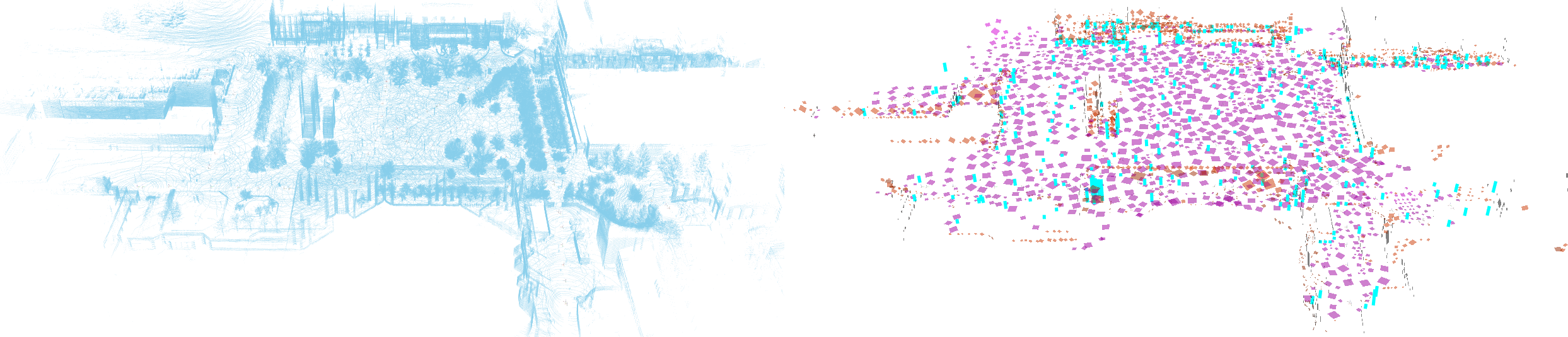}
  }
  \subfigure[HeLiPR DCC Region 05]{
    \includegraphics[width=0.48\linewidth]{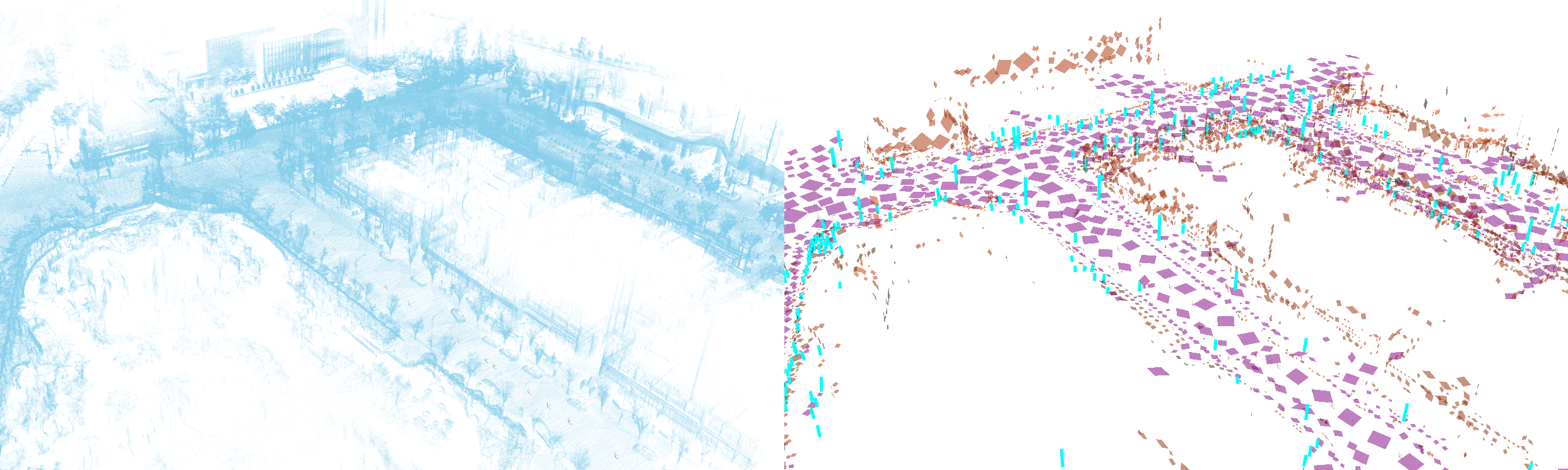}
  }
  \subfigure[HeLiPR KAIST Region 00]{
    \includegraphics[width=0.48\linewidth]{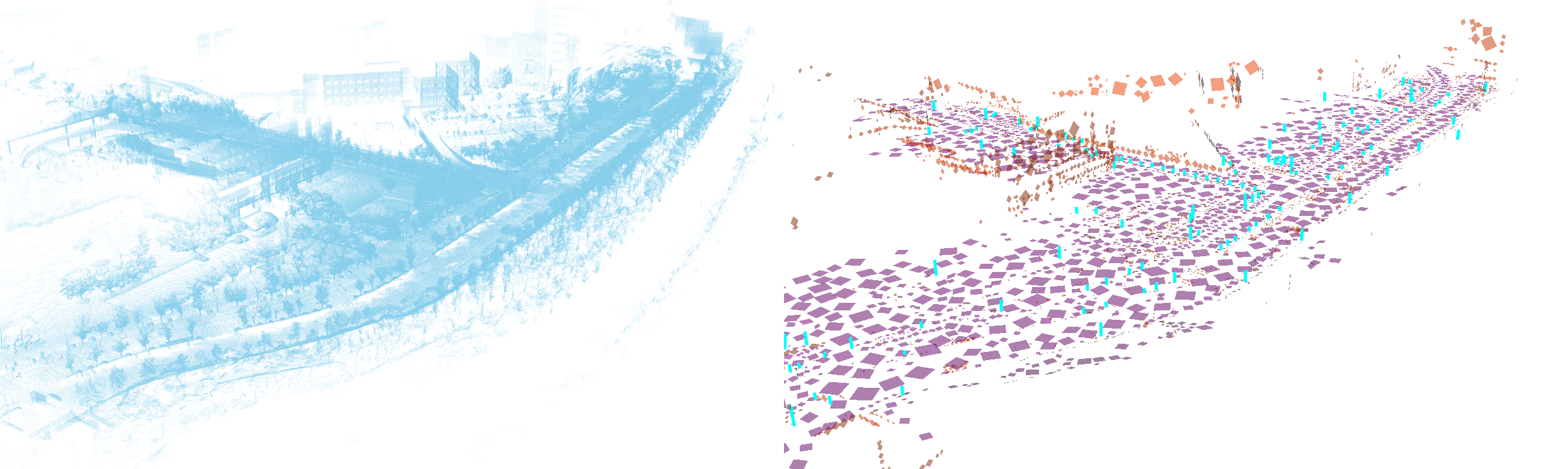}
  }
  \subfigure[HeLiPR Town Region 05]{
    \includegraphics[width=0.48\linewidth]{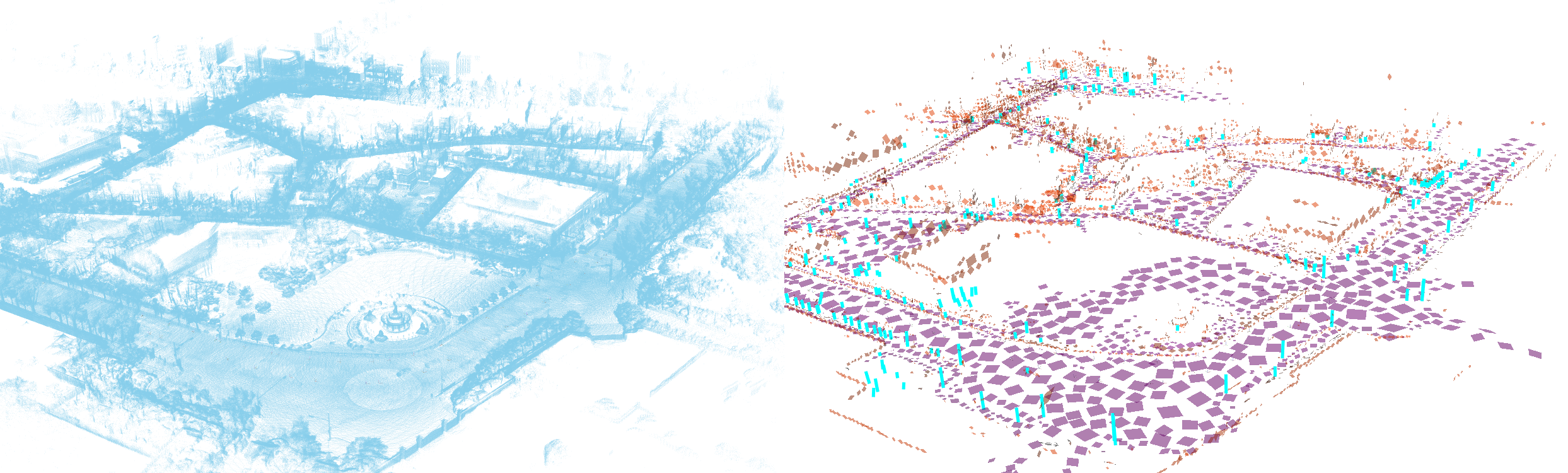}
  }
  \caption{Visualization of maps on NCLT and HeLiPR. We present the downsampled point cloud maps and the SLIM-generated maps from the same perspective view. The point cloud maps are built by accumulating LiDAR points on the robot poses estimated by SLIM. SLIM provides line and plane-based maps that are with high consistency (Section~\ref{sec:accuracy}) and can be reused for robot localization (Section~\ref{sec:localization}). More importantly, the map is much more lightweight (Section~\ref{sec:lightweight}) and scalable for long-term use (Section~\ref{sec:scalability}).}
  \label{fig:mapping_visualization}
\end{figure*}

%% file: figure-caption/trajectory_accuracy.tex
\begin{figure*}[t]
  \centering
  \subfigure[Roundabout Region 00]{
    \includegraphics[width=0.23\linewidth]{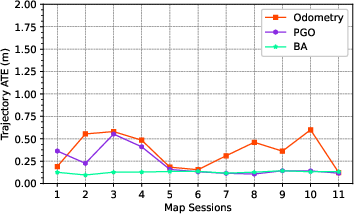}
  }
  \subfigure[Roundabout Region 01]{
    \includegraphics[width=0.23\linewidth]{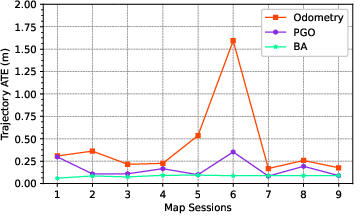}
  }
  \subfigure[Town Region 00]{
    \includegraphics[width=0.23\linewidth]{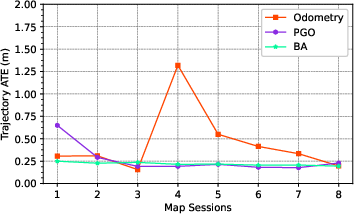}
  }
  \subfigure[Town Region 01]{
    \includegraphics[width=0.23\linewidth]{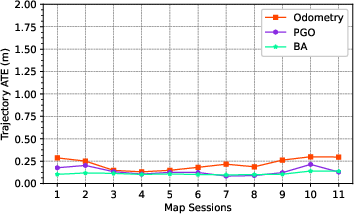}
  }
  \subfigure[DCC Region 00]{
    \includegraphics[width=0.23\linewidth]{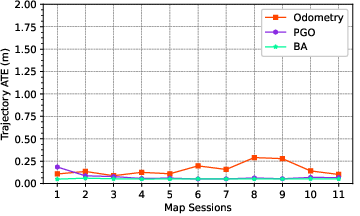}
  }
  \subfigure[DCC Region 01]{
    \includegraphics[width=0.23\linewidth]{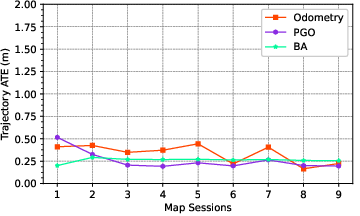}
  }
  \subfigure[KAIST Region 00]{
    \includegraphics[width=0.23\linewidth]{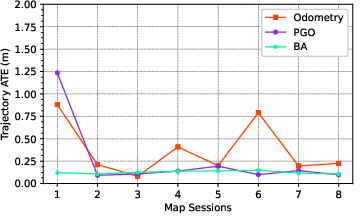}
  }
  \subfigure[KAIST Region 01]{
    \includegraphics[width=0.23\linewidth]{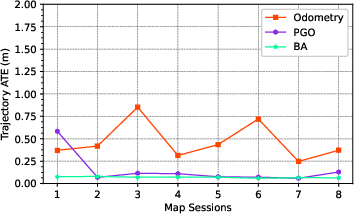}
  }
  \caption{Quantitative results of mapping accuracy on HeLiPR. The evaluation metric is the RMSE of absolute trajectory error. The overall results indicate that: (1) PGO (purple) and BA (cyan) effectively reduce errors and improve the consistency of the global map. (2) In certain cases, inaccurate submaps degrade the refinement results of PGO, while BA can almost entirely mitigate such impacts, demonstrating the necessity and effectiveness of our proposed BA.}

  \label{fig:trajectory_accuracy}
\end{figure*}

%% file: figure-caption/M2DGR.tex
\begin{figure}[t]
	\centering
          \subfigure[Multi-session map on M2DGR]{
            \includegraphics[width=0.46\linewidth]{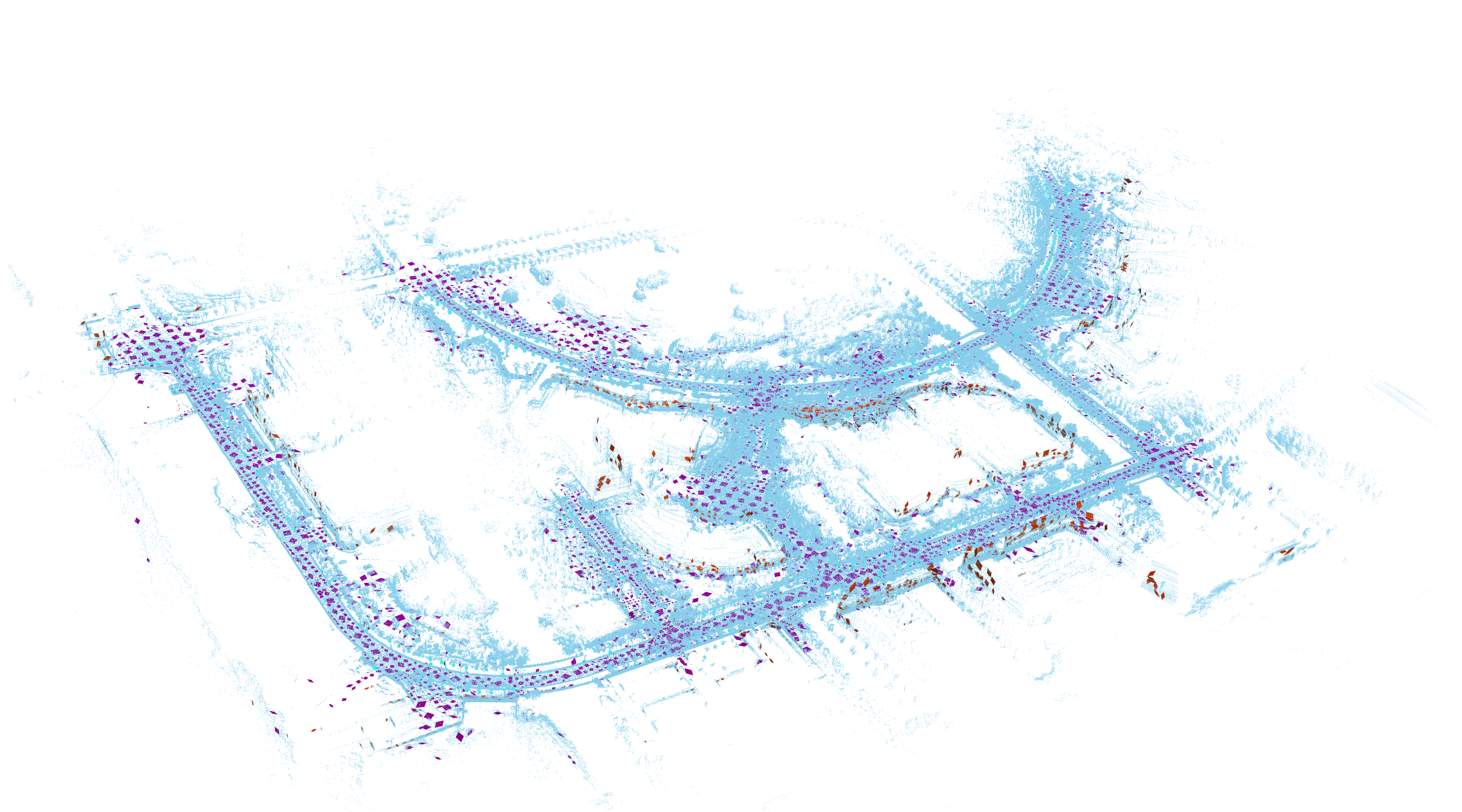}
          }
    \subfigure[Map view at the campus gate]{
            \includegraphics[width=0.46\linewidth]{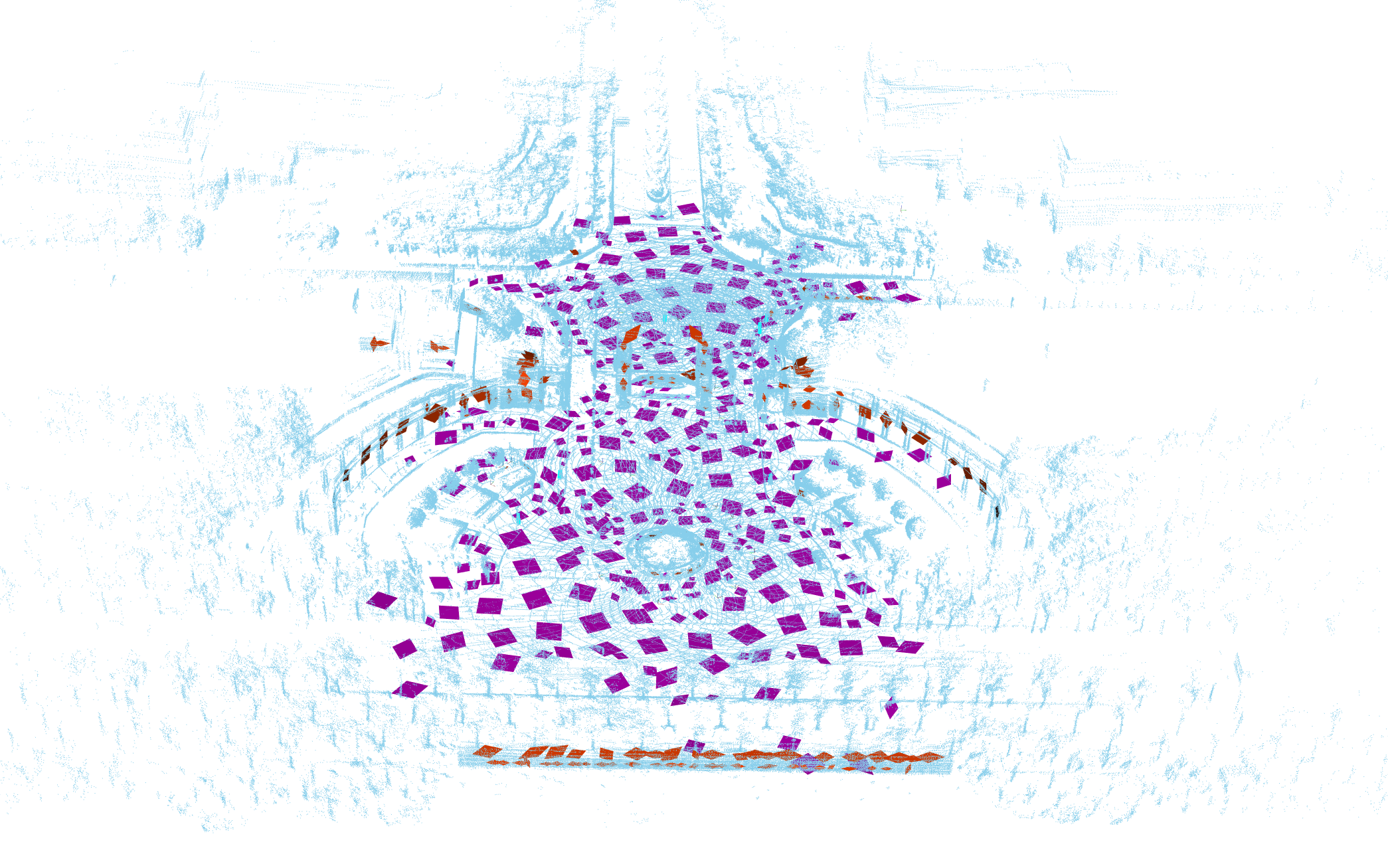}
          }
	\caption{Demonstration in a university campus on M2DGR. For comparison, the point cloud maps are built by accumulating LiDAR points based on the robot poses estimated by SLIM.}
	\label{figure:m2dgr}
\end{figure}

%% file: figure-caption/cloud_compare.tex

\begin{figure}[t]
  \centering
  \subfigure[DCC Region 06]{
    \includegraphics[width=0.46\linewidth]{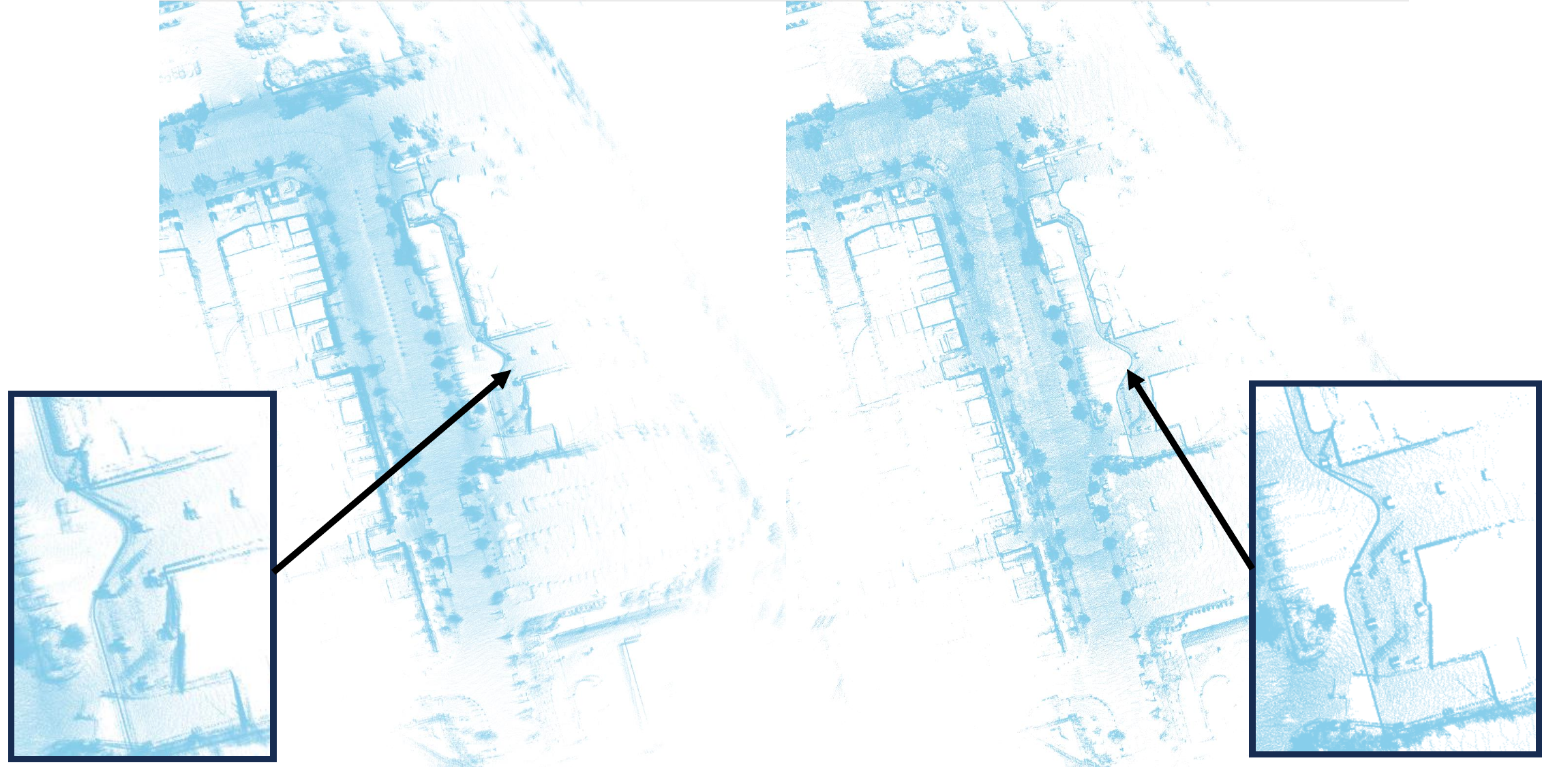}
  }
  \subfigure[Roundabout Region 00]{
    \includegraphics[width=0.46\linewidth]{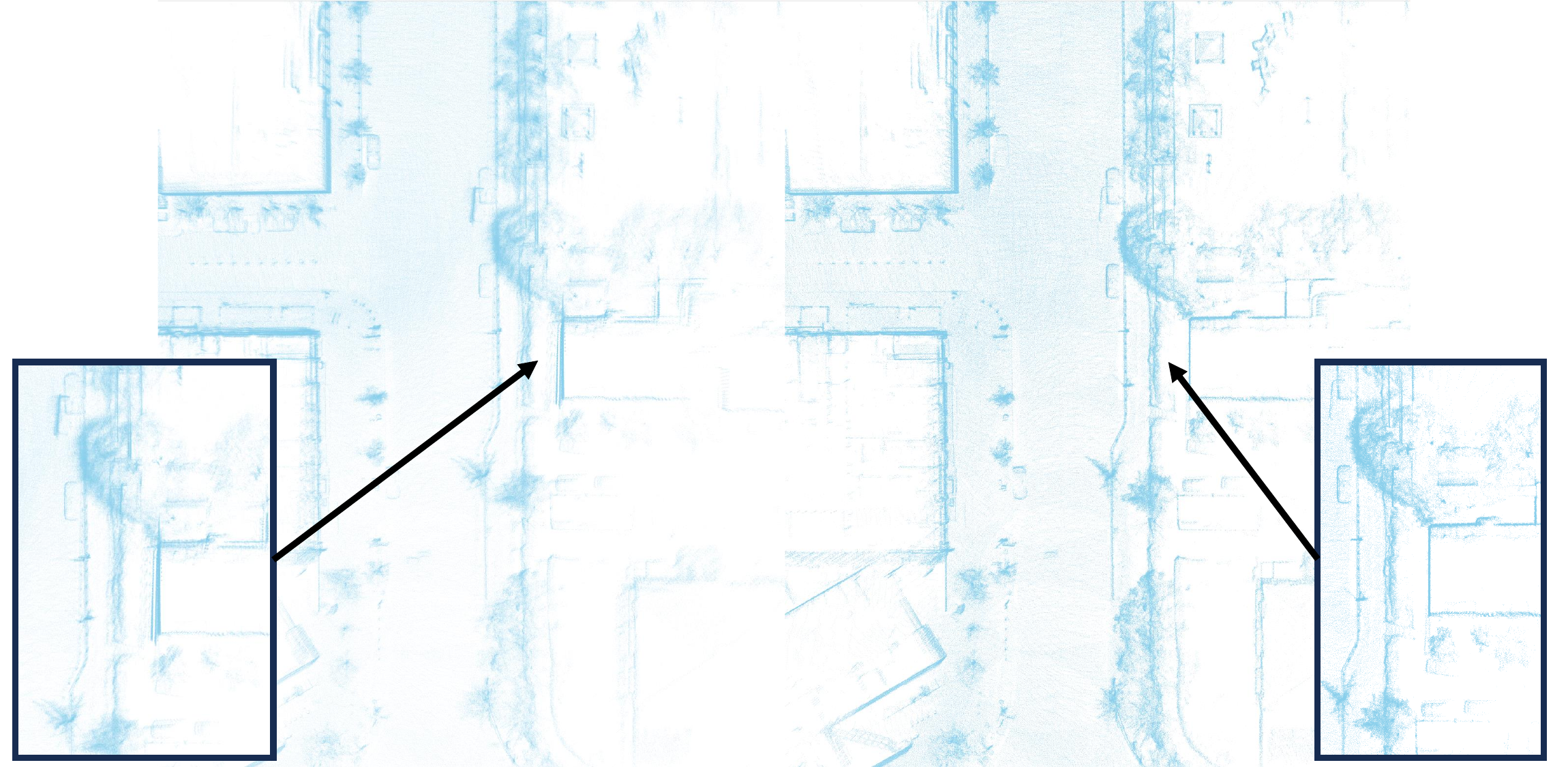}
  }
  \caption{Visualizing the BA effectiveness in HeLiPR. The point cloud maps are built by accumulating LiDAR points on robot poses provided by SLIM. The poses on the left side are obtained before the LiDAR BA, i.e., only using the PGO. After applying the BA, the poses on the right have fewer drifts, resulting in high-precision maps in zoomed-in views. This figure demonstrates the effectiveness of the proposed LiDAR BA on parameterized representations.}
  \label{fig:ba_cloud_compare}
\end{figure}

%% file: table-caption/storage.tex
\begin{table}[!t]
    \centering
    \caption{Memory Consumption (Mega Byte, MB) on KITTI }
    \renewcommand\arraystretch{1.2}
    \resizebox{0.48\textwidth}{!}{
    \begin{tabular}{cccc}
    \hline
    \hline
    Sequence (Length) & 00 (3.7 Km) & 05 (2.2 Km)& 08 (3.2 Km) \\
    \hline
    Raw point cloud & 13624.2& 8284.7& 12214.1 \\
    Point cloud ($r$=0.1m) & 2831.46 & 1852.96 & 2769.22 \\
    Point cloud ($r$=0.3m) & 52.03 & 32.58 & 70.41 \\
    Surfel map~\cite{behley2018efficient} & 887.7& 512.6& 835.7 \\
    Mesh map~\cite{vizzo2021poisson} & 2032.9& 1317.4& 1894.1 \\
    VDB TSDF map~\cite{vizzo2022vdbfusion} & 748.1& 434.6 & 958.6 \\
    SHINE map~\cite{zhong2023shine} & 160.6& 114.2& 189.9 \\
    LONER\footnotemark~\cite{isaacson2023loner} & 150.1 & 150.1 & 150.1 \\
    NeRF-LOAM\footnotemark~\cite{deng2023nerf} & 953.61 & 800.69 & 1424.9 \\ 
    PIN-SLAM~\cite{pan2024pin} & 102.1& 66.3 & 138.8 \\
    \hline
    SLIM & \textbf{12.9} & \textbf{7.2} & \textbf{9.2} \\
    SLIM (L) & \textbf{0.5} & \textbf{0.3} & \textbf{0.4} \\
    \hline
    \hline
    \end{tabular}
    }
    \label{tab:storage}
\end{table}

\footnotetext[1]{LONER fails to run on certain KITTI sequences, resulting in large drifts. We use $\times$ to indicate these ``Failed'' sequences. As stated in \cite{isaacson2023loner}, LONER needs further revisions to operate city-scale scenarios.}
\footnotetext[2]{LONER has the same memory consumption across the sequences due to its map structure. The adaption of LONER to KITTI is available at \href{https://github.com/qiaozhijian/LONER-KITTI}{https://github.com/qiaozhijian/LONER-KITTI}.}

\footnotetext[3]{Due to the limitations of GPU memory, NeRF-LOAM is performed on segments of KITTI sequences.}

%% file: figure-caption/nclt_storage.tex
\begin{figure*}[t]
  \centering
  \subfigure[NCLT Region 00]{
    \includegraphics[width=0.23\linewidth]{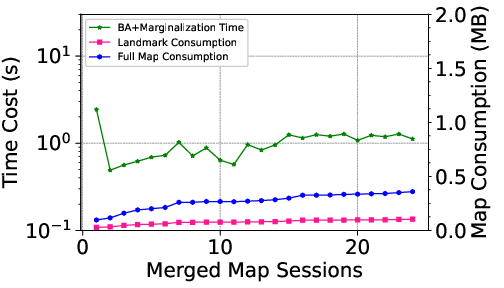}
  }
  \subfigure[NCLT Region 01]{
    \includegraphics[width=0.23\linewidth]{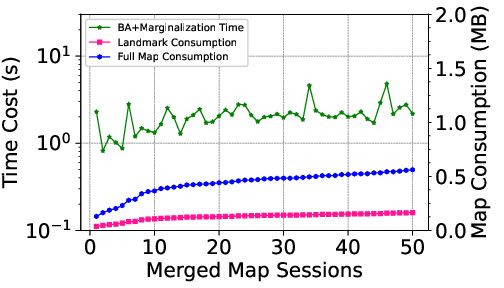}
  }
  \subfigure[NCLT Region 02]{
    \includegraphics[width=0.23\linewidth]{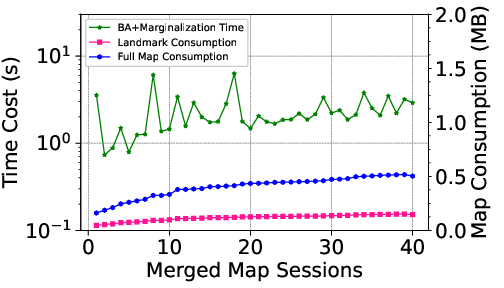}
  }
  \subfigure[NCLT Region 03]{
    \includegraphics[width=0.23\linewidth]{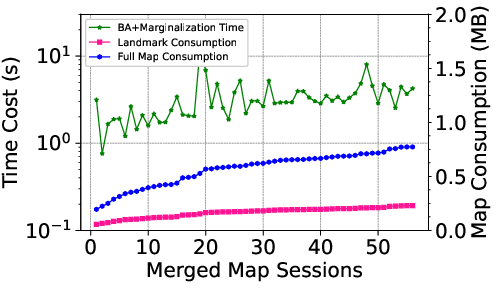}
  }
  \caption{Quantitative results of map consumption of SLIM and the time cost for map refinement. The green line represents ``BA+Marginalization Time''; the pink line is ``Landmark Consumption''; and the blue line is ``Full Map Consumption''. The results on NCLT indicate that (1) the provided map is lightweight and only uses low storage consumption; (2) the consumption and time cost remain stable as the number of sessions increases. }
  \label{fig:nclt_storage}
\end{figure*}

%% file: figure-caption/nfr_compare.tex
\begin{figure*}[t]
  \centering
  \subfigure[Roundabout Region 00]{
    \includegraphics[width=0.23\linewidth]{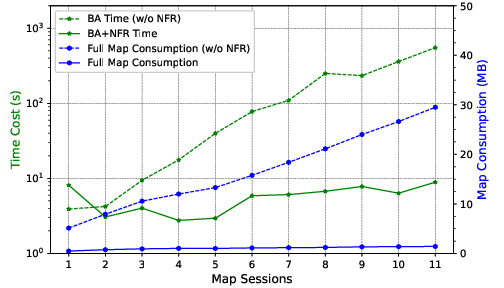}
  }
  \subfigure[Town Region 00]{
    \includegraphics[width=0.23\linewidth]{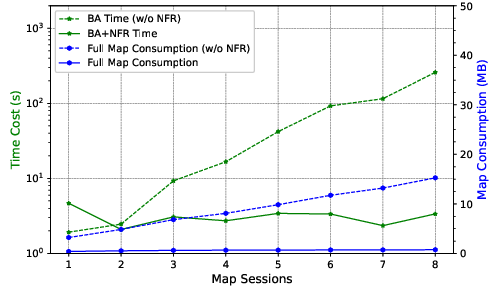}
  }
  \subfigure[DCC Region 00]{
    \includegraphics[width=0.23\linewidth]{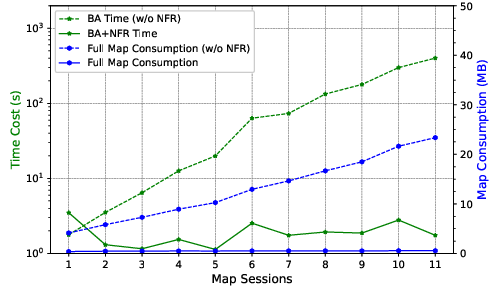}
  }
  \subfigure[KAIST Region 00]{
    \includegraphics[width=0.23\linewidth]{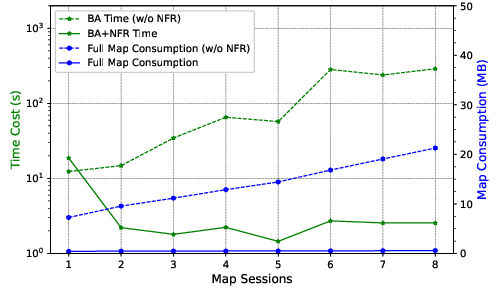}
  }
  \caption{Quantitative results of using and without using marginalization on HeLiPR. The blue solid line represents ``Full Map Consumption''; the blue dashed line is ``Full Map Consumption (w/o Marginalization)''; the green solid line represents ``BA+Marginalization Time'' and the green dashed line is ``BA Time (w/o Marginalization)''. If the marginalization is enabled, the time cost and map consumption are bounded. More specifically, the BA and marginalization cost less than 10 seconds. The map consumes less than 1 MB for one region.}
  \label{fig:nfr_compare}
\end{figure*}

%% file: figure-caption/nfr_prior_compare.tex
\begin{figure*}[t]
  \centering
  \subfigure[DCC]{
    \includegraphics[width=0.22\linewidth]{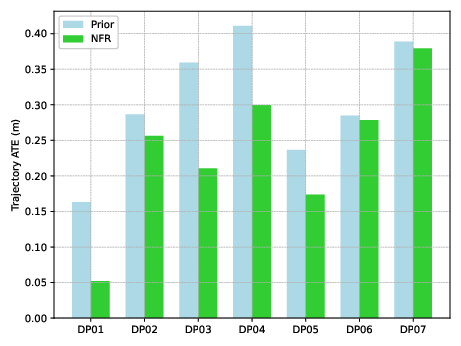}
  }
  \subfigure[KAIST]{
    \includegraphics[width=0.22\linewidth]{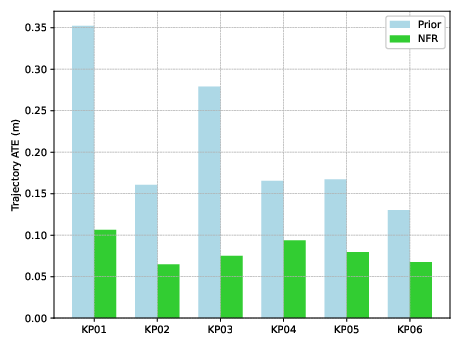}
  }
  \subfigure[Roundabout]{
    \includegraphics[width=0.22\linewidth]{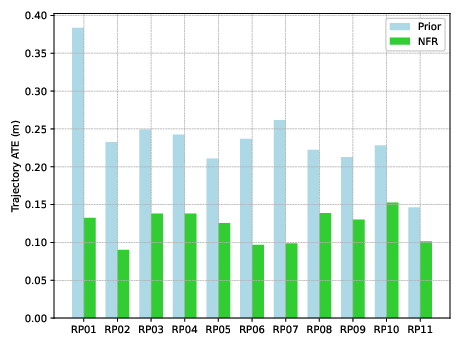}
  }
  \subfigure[Town]{
    \includegraphics[width=0.22\linewidth]{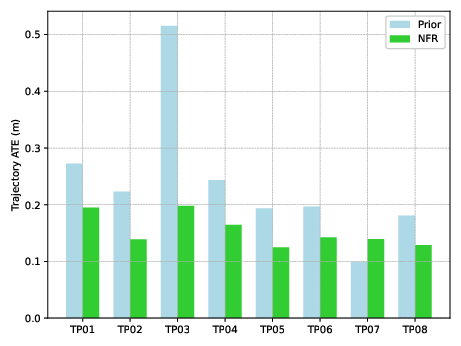}
  }
  \caption{Quantatitive results of using and without using NFR on mapping accuracy, i.e., the absolute trajectory error. The results indicate that our proposed map-entric NFR performs better than using the prior information matrix, thus demonstrating its effectiveness for the marginalization of long-term mapping.
  }
  \label{fig:nfr_prior_compare}
\end{figure*}

%% file: figure-caption/reduced_graph.tex
\begin{figure}[t]
	\centering
          \includegraphics[width=0.85\linewidth]{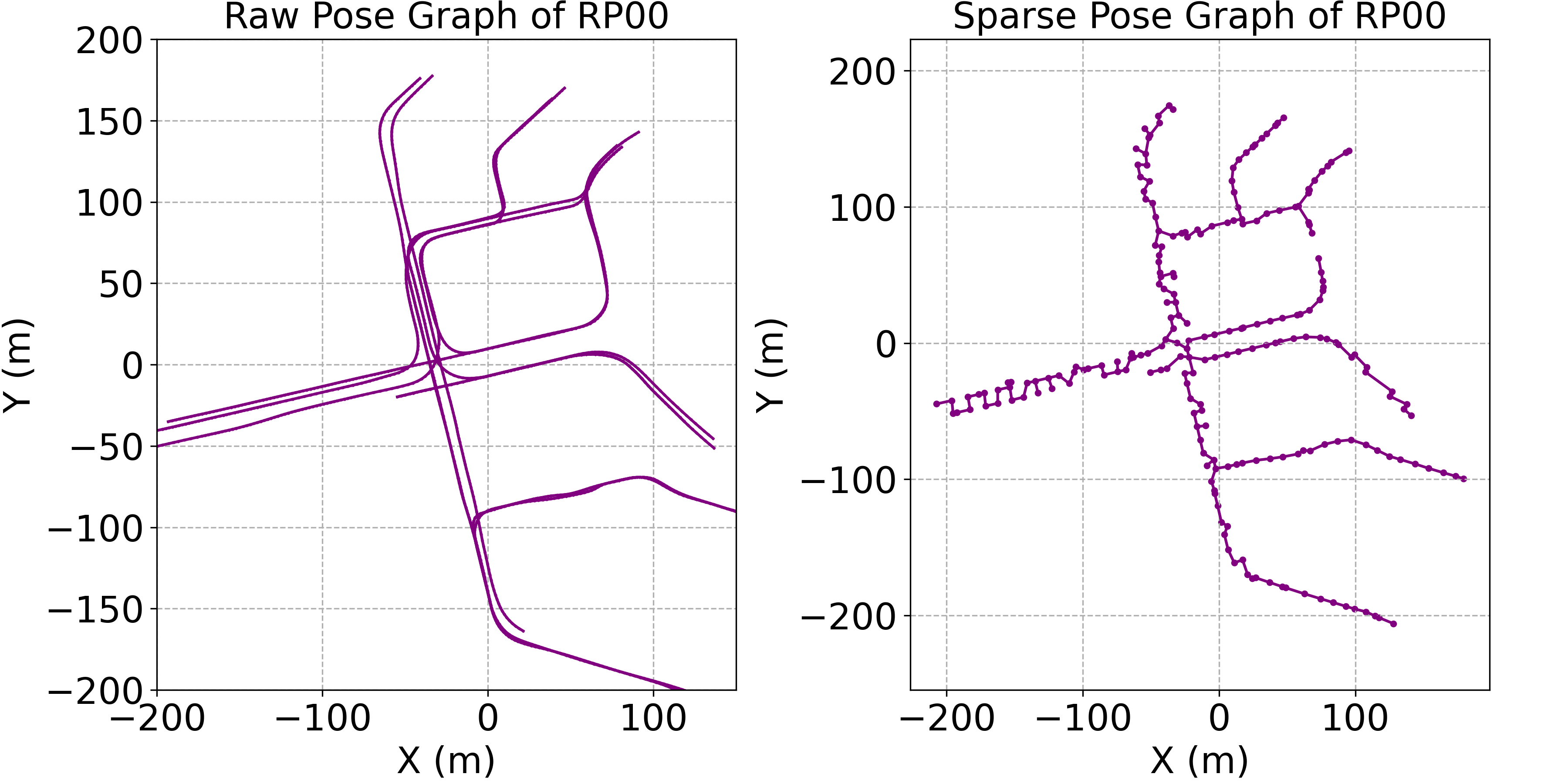}
	\caption{The reduced sparse pose graph after applying marginalization on HeLiPR. The multi-session data results in multiple pose graphs in the same region. The marginalization provides not only a more compact topological structure but also the information matrix for long-term mapping.
 }
	\label{figure:reduced_graph}
\end{figure}

%% file: table-caption/localization.tex
\begin{table}[t]
    \centering
    \renewcommand\arraystretch{1.2}
    \caption{Online Localization on SLIM-generated Maps: Absolute Trajectory Error (RMSE, Centimeters) }
    \resizebox{0.48\textwidth}{!}{
    \begin{tabular}{ccccccccc}
    \hline
    \hline
    Region & DP01 & DP02 & DP03 & DP04 & DP05 & DP06 & DP07 & KP01\\
    ATE & 2.8 & 8.8 & 5.8 & 32.6 & 6.2 & 8.5 & 17.9 & 7.9\\
    \hline
    Region & KP02 & KP03 & KP04 & KP05 & KP06 & RP01 & RP02 & RP03\\
    ATE & 5.9 & 6.9 & 8.6 & 3.3 & 5.0 & 8.4 & 5.5 & 6.4\\
    \hline
    Region & RP04 & RP05 & RP06 & RP07 & RP08 & RP09 & RP10 & RP11\\
    ATE & 9.0 & 7.1 & 5.4 & 5.1 & 6.9 & 11.5 & 9.2 & 8.6\\
    \hline
    Region & TP01 & TP02 & TP03 & TP04 & TP05 & TP06 & TP07 & TP08\\
    ATE & 19.0 & 11.1 & 35.8 & 6.5 & 9.2 & 9.2 & 12.2 & 14.2\\
    \hline
    \hline
    \end{tabular}
    }
    \label{tab:localization}
\end{table}

%% file: sections/limitations.tex
\section{Discussions}
\label{sec:limitations}

Though the designed SLIM system is versatile for long-term LiDAR mapping, potential limitations exist for practical applications. The two key limitations are listed as follows:
\begin{enumerate}
  \item A closed-form solution for the NFR requires that the dimension of the recovered residual matches the dimension of the state variables. This requires the Jacobian matrix must be square. In SLAM backends, this condition is not always met. For example, incorporating GPS constraints, pitch-roll constraints, or other constraints may result in mismatched dimensions between residuals and state variables. A crucial factor in our system is that the dimension of lines and planes precisely matches their residual dimension, easily satisfying the condition for a closed-form solution. However, if more information is fused, this condition may not hold.
  \item Although an iterative solution for NFR is available~\cite{mazuran2016nonlinear}, the computational cost increases rapidly for extremely large-scale maps. Fortunately, for maps containing between $1.0 \times 10^4$ and $1.0 \times 10^5$ landmarks, the proposed SLIM can achieve a single-step solution within a few seconds, which is acceptable for iterative methods. If more map landmarks are involved, the SLIM system requires more time to provide the solution.
\end{enumerate}

The SLIM backend remains effective for visual SLAM, as its problem formulation is inherently aligned with the principles of visual bundle adjustment. For instance, in the context of a crowdsourced semantic mapping task on roads, the line and plane features within the framework can be effortlessly replaced with semantic entities such as poles, lane markings, and traffic signs. By incorporating updated landmark definitions and residual formulations, the backend solver retains its adaptability to novel problem representations, thereby ensuring robust and precise solutions.

Despite the focus on urban environments, the SLIM system could also be applied to other structured environments, such as forests and indoors. In forests, the ground and trees can be parameterized into planes and lines. Indoor environments are highly structured with elements like walls and columns, which are typical planes and lines. We also want to emphasize that the goal of the SLIM system is not to achieve state-of-the-art performance. This study primarily aims to provide insights into lightweightness and scalability for LiDAR mapping in urban environments, which are the motivations as stated in Section~\ref{sec:introduction}.


%% file: sections/conclusion.tex
\section{Conclusion and Future Study}
\label{sec:conclusion}




In this study, we present SLIM, a lightweight and scalable LiDAR mapping system for urban environments. Our contributions span from front-end feature extraction to back-end map optimization and management. Experiments conducted on various datasets with different LiDAR sensor types demonstrate the effectiveness of the SLIM system. The results show that SLIM can merge, refine, and maintain new maps using only line and plane representations.

Future studies could explore several promising directions. First, SLIM could be deployed in other structured environments similar to urban scenarios. At the front end, learning-based feature extraction could reduce the need for parameter tuning. At the back end, experiments currently rely on manual region divisions, and a more adaptive approach could be a better choice for long-term autonomy.

%% file: sections/appendix.tex
\appendix
\label{sec:appendix}

The derivation of map refinement parts is detailed in this section. The first part is the residual and Jacobian matrices with respect to line landmarks and robot poses. As described in Section~\ref{sec:representations}, the line landmark is parameterized by two rotation angles (roll and pitch) and two translations:
\begin{equation} \small
\begin{aligned}
    \mathbf{l}_{\mathcal{L}}&=f\left( \mathbf{n},\mathbf{c} \right) =g\left( \alpha ,\beta ,x,y \right)   \\
    \mathbf{n}&=\mathbf{R}\left( \alpha ,\beta \right) \mathbf{u}_z
    \\
    \mathbf{q}&=\mathbf{R}\left( \alpha ,\beta \right) \left( \mathbf{u}_xx+\mathbf{u}_yy \right) \\
\end{aligned}
\label{eq:lineLMDef}
\end{equation}
\par Theoretically, the degrees of freedom for point-to-line residual is 2 because the constraints along the direction of the line are redundant. Thus, we project the point-to-line vector to the original orthogonal frame based on the definition of line landmark, described as follows:
\begin{equation} \small
\begin{aligned}
\mathbf{r}_{\mathcal{L}}\left( \mathbf{R}_f^w,\mathbf{p}_f^w,\mathbf{l}_{\mathcal{L}} \right) =\mathbf{R}^T\left( \alpha ,\beta \right) \left( \mathbf{I}_3-\mathbf{nn}^T \right) \left( \mathbf{p}^w-\mathbf{q} \right) 
\end{aligned}
\end{equation}
where $\mathbf{p}^w=\mathbf{R}_f^w\mathbf{\hat p}+\mathbf{p}_f^w$ and $\mathbf{\hat p}$ is the observation points in local keyframe coordinate. Note that the third dimension of this residual is redundant; we will now derive its equivalent form. From Equation~\eqref{eq:Rotation2dof} and \eqref{eq:lineLMDef}, we can derive the following equation:
\begin{equation} \small
\begin{aligned}
&\mathbf{R}^T\left( \alpha ,\beta \right) \left( \mathbf{I}_3-\mathbf{nn}^T \right) \\
&=\mathbf{R}^T\left( \alpha ,\beta \right) \left( \mathbf{I}-\mathbf{R}\left( \alpha ,\beta \right) \mathbf{u}_z\mathbf{u}_{z}^{T}\mathbf{R}^T\left( \alpha ,\beta \right) \right) 
\\
&=\left( \mathbf{R}^T\left( \alpha ,\beta \right) -\mathbf{u}_z\mathbf{u}_{z}^{T}\mathbf{R}^T\left( \alpha ,\beta \right) \right)\\
&=\left( \mathbf{I}_3-\mathbf{u}_z\mathbf{u}_{z}^{T} \right) \mathbf{R}^T\left( \alpha ,\beta \right) 
\end{aligned}   
\end{equation}
\par Therefore, the equivalent form of the point-to-line residual can be transformed into the subsequent equation:
\begin{equation} \small
\begin{aligned}
&\mathbf{r}_{\mathcal{L}}\left( \mathbf{R}_f^w,\mathbf{p}_f^w,\mathbf{l}_{\mathcal{L}} \right) =\left( \mathbf{I}_3-\mathbf{u}_z\mathbf{u}_{z}^{T} \right) \mathbf{R}^T\left( \alpha ,\beta \right) \left( \mathbf{p}^w-\mathbf{q} \right) 
\\
&=\left( \mathbf{I}-\mathbf{u}_z\mathbf{u}_{z}^{T} \right) \left( \mathbf{R}^T\left( \alpha ,\beta \right) \mathbf{p}^w-\left( \mathbf{u}_xx+\mathbf{u}_yy \right) \right) 
\\
&=\mathbf{R}^T\left( \alpha ,\beta \right) _{\left[ 2\times 3 \right]}\mathbf{p}^w-\left( \mathbf{u}_xx+\mathbf{u}_yy \right) _{\left[ 2\times 1 \right]}
\end{aligned}
\end{equation}
\par Then we can obtain the Jacobian matrix:
\begin{equation} \small
\begin{aligned}
\mathbf{R}_{\alpha}&=
\begin{bmatrix}
    0&		\cos \alpha \sin \beta&		-\sin \alpha \sin \beta\\
    0&		-\sin \alpha&		-\cos \alpha\\
\end{bmatrix} \\
\mathbf{R}_{\beta}&=
\begin{bmatrix}
    -\sin \beta&		\sin \alpha \cos \beta&		\cos \alpha \cos\beta\\
    0&		0&		0\\
\end{bmatrix} \\
\frac{\partial \mathbf{r}_{\mathcal{L}}}{\partial \alpha}&=\mathbf{R}_{\alpha}\mathbf{p}^w, \qquad
\frac{\partial \mathbf{r}_{\mathcal{L}}}{\partial \beta}=\mathbf{R}_{\beta}\mathbf{p}^w\\
\frac{\partial \mathbf{r}_{\mathcal{L}}}{\partial \left( x, y \right)}&=-\mathbf{I}_2, \qquad \frac{\partial \mathbf{r}_{\mathcal{L}}}{\partial \mathbf{p}_f^w}=\mathbf{R}^T\left( \alpha ,\beta \right) _{\left[ 2\times 3 \right]} \\
\frac{\partial \mathbf{r}_{\mathcal{L}}}{\partial \delta \theta _f^w}&=-\mathbf{R}^T\left( \alpha ,\beta \right) _{\left[ 2\times 3 \right]}\left( \mathbf{R}_f^w \right) ^T\left[ \mathbf{\hat p} \right] _{\times}
\end{aligned}
\label{eq:lineJacobian}
\end{equation}


The second part is the residual and Jacobian matrices with respect to plane landmarks and robot poses. The plane landmark is modeled by two rotation angles (roll and pitch) and one translation, described as follows: 
\begin{equation} \small
    \mathbf{l}_{\mathcal{S}}=f\left( \mathbf{n},d \right) =g\left( \alpha ,\beta ,d \right)     
\end{equation}
\par The point-to-plane residual is defined by the point-to-plane distance as follows:
\begin{equation} \small
\mathbf{r}_{\mathcal{S}}\left( \mathbf{R}_f^w,\mathbf{p}_f^w,\mathbf{l}_{\mathcal{S}} \right) =\mathbf{n}^T\mathbf{p}^w+d
\end{equation}
and the Jacobian matrices can be derived as

\begin{equation} \small
\begin{aligned}
\frac{\partial \mathbf{r}_{\mathcal{S}}}{\partial \alpha}&=\left( \mathbf{p}^w \right) ^T\left[ \begin{array}{c}
	0\\
	\cos \alpha \cos \beta\\
	-\sin \alpha \cos \beta\\
\end{array} \right] 
\\
\frac{\partial \mathbf{r}_{\mathcal{S}}}{\partial \beta}&=\left( \mathbf{p}^w \right) ^T\left[ \begin{array}{c}
	-\cos \beta\\
	-\sin \alpha \sin \beta\\
	-\cos \alpha \sin \beta\\
\end{array} \right] 
\\
\frac{\partial \mathbf{r}_{\mathcal{S}}}{\partial d}&=1
, \qquad \frac{\partial \mathbf{r}_{\mathcal{S}}}{\partial \mathbf{p}_f^w}=\mathbf{n}^T \\
\frac{\partial \mathbf{r}_{\mathcal{S}}}{\partial \delta \theta _f^w}&=-\mathbf{n}^T\left( \mathbf{R}_f^w \right) ^T\left[ \mathbf{\hat p} \right] _{\times}
\end{aligned}
\end{equation}


To this end, the Jacobian matrices for map refinement (Section~\ref{sec:refinement}) are obtained under the scheme of vectorized representations (Section~\ref{sec:representations}). The errors of maps are reduced by applying the map refinements, as presented in Table~\ref{tab:kitti_acc} and Figure~\ref{fig:trajectory_accuracy}, thus validating the effectiveness of the derivations in the appendix.